\documentclass{article}
\usepackage[preprint]{neurips_2025}

\usepackage{tikz}
\usetikzlibrary{positioning, calc, fit, backgrounds}

\newcommand{\eglist}[1]{\textit{e.g.,} #1}
\definecolor{mainblue}{HTML}{C9DAF8}
\definecolor{lightblue}{HTML}{D9EAD3}
\definecolor{mainred}{HTML}{F4CCCC}
\definecolor{mainpurple}{HTML}{D9D2E9}

\tikzset{
    box/.style={draw, rounded corners=2pt, align=left, minimum height=1em, inner sep=1.5pt},
    mainbox/.style={fill=#1, rounded corners=3pt, inner ysep=2pt, inner xsep=2pt},
    subbox/.style={box, fill=#1, font=\bfseries\scriptsize, text width=0.15\textwidth, align=center},
    methodbox/.style={box, fill=#1!50!white, font=\scriptsize, text width=0.2\textwidth},
    egbox/.style={box, fill=#1!50!white, font=\scriptsize, text width=0.56\textwidth}
}

\usepackage[utf8]{inputenc} 
\usepackage[T1]{fontenc}    
\usepackage{hyperref}       
\usepackage{url}            
\usepackage{booktabs}       
\usepackage{amsfonts}       
\usepackage{nicefrac}       
\usepackage{microtype}      
\usepackage{xcolor}         
\usepackage{subcaption}

\usepackage[most]{tcolorbox}
\tcbuselibrary{skins,breakable}
\usepackage{hyperref}
\usepackage{array}
\usepackage{microtype}
\usepackage{graphicx}
\usepackage{booktabs} 
\usepackage{xcolor}
\usepackage{tabularray}

\definecolor{lightgray}{HTML}{F2F2F2}
\definecolor{lightblue}{HTML}{EAF2F8}
\definecolor{lightgreen}{HTML}{E8F6F3}
\definecolor{headergray}{gray}{0.9}
\definecolor{rowblue}{RGB}{230, 240, 255}
\definecolor{rowgreen}{RGB}{230, 255, 240}
\definecolor{rowyellow}{RGB}{255, 255, 230}
\definecolor{roworange}{RGB}{255, 245, 230}
\definecolor{rowpurple}{RGB}{245, 230, 255}
\usepackage{amsmath}
\usepackage{amssymb}
\usepackage{mathtools}
\usepackage{amsthm}
\usepackage{colortbl}
\usepackage[capitalize,noabbrev]{cleveref}

\NewDocumentCommand{\shuiwang}
{ mO{} }{\textcolor{blue}{\textsuperscript{\textit{Shuiwang Ji}}\textsf{\textbf{\small[#1]}}}}

\usepackage[textsize=tiny]{todonotes}
\NewDocumentCommand{\heng}
{ mO{} }{\textcolor{red}{\textsuperscript{\textit{Heng}}\textsf{\textbf{\small[#1]}}}}

\NewDocumentCommand{\xiaofeng}
{ mO{} }{\textcolor{red}{\textsuperscript{\textit{Xiaofeng}}\textsf{\textbf{\small[#1]}}}}

\title{Autonomous Agents for Scientific Discovery: Orchestrating Scientists, Language, Code, and Physics}

\author{%
\textbf{Lianhao Zhou}\textsuperscript{$1$}\quad
\textbf{Hongyi Ling}\textsuperscript{$1$}\quad
\textbf{Cong Fu}\textsuperscript{$1$}\quad
\textbf{Yepeng Huang}\textsuperscript{$2$}\quad \\
\textbf{Michael Sun}\textsuperscript{$3$}\quad
\textbf{Wendi Yu}\textsuperscript{$1$}\quad
\textbf{Xiaoxuan Wang}\textsuperscript{$4$}\quad
\textbf{Xiner Li}\textsuperscript{$1$}\quad \\
\textbf{Xingyu Su}\textsuperscript{$1$}\quad 
\textbf{Junkai Zhang}\textsuperscript{$4$}\quad
\textbf{Xiusi Chen}\textsuperscript{$5$}\quad
\textbf{Chenxing Liang}\textsuperscript{$1$}\quad \\
\textbf{Xiaofeng Qian}\textsuperscript{$6,7,8$}\quad
\textbf{Heng Ji}\textsuperscript{$5$}\quad
\textbf{Wei Wang}\textsuperscript{$4$}\quad
\textbf{Marinka Zitnik}\textsuperscript{$2$}\quad 
\textbf{Shuiwang Ji}\textsuperscript{$1,6,9$}\thanks{Correspondence to: Shuiwang Ji <sji@tamu.edu>}\\
\textsuperscript{$1$}Department of Computer Science and Engineering, Texas A\&M University  \\ 
\textsuperscript{$2$}Department of Biomedical Informatics, Harvard Medical School\\
\textsuperscript{$3$}Computer Science and Artificial Intelligence Laboratory, Massachusetts Institute of Technology\\
\textsuperscript{$4$}Department of Computer Science, University of California, Los Angeles\\
\textsuperscript{$5$}Siebel School of Computing and Data Science, University of Illinois Urbana Champaign\\
\textsuperscript{$6$}Department of Materials Science and Engineering, Texas A\&M University  \\
\textsuperscript{$7$}Department of Electrical and Computer Engineering, Texas A\&M University\\
\textsuperscript{$8$}Department of Physics and Astronomy, Texas A\&M University\\
\textsuperscript{$9$}J. Mike Walker '66 Department of Mechanical Engineering, Texas A\&M University\\
}   

\begin{document}
\maketitle

\begin{abstract}
Computing has long served as a cornerstone of scientific discovery. Recently, a paradigm shift has emerged with the rise of large language models (LLMs), introducing autonomous systems, referred to as agents, that accelerate discovery across varying levels of autonomy.
These language agents provide a flexible and versatile framework that orchestrates interactions with human scientists,
natural language, computer language and code, and physics.
This paper presents our view and vision of LLM-based scientific agents and their growing role in transforming the scientific discovery lifecycle, from hypothesis discovery, experimental design and execution, to result analysis and refinement. We critically examine current methodologies, emphasizing key innovations, practical achievements, and outstanding limitations. Additionally, we identify open research challenges and outline promising directions for building more robust, generalizable, and adaptive scientific agents. Our analysis highlights the transformative potential of autonomous agents to accelerate scientific discovery across diverse domains.
\end{abstract}

\section{Introduction}

Scientific discovery is fundamental to advancing human knowledge, driving innovations across diverse fields such as biology, chemistry, physics, and materials science, etc.~\citep{zhang2023artificial,wang2023scientific}. 
The pivotal role of computing in advancing discoveries across the natural sciences has long been recognized~\citep{dirac1929quantum}, with impacts ranging from quantum mechanics~\citep{landau2013quantum,Tong2025QuantumMechanics} to fluid dynamics~\citep{landau2013fluid,Tong2025FluidMechanics}. 
For many years, the process of scientific discovery has been predominantly dependent on human intuition, expertise, and iterative experimentation~\citep{xu2023ai}. Nevertheless, human-driven discovery processes sometimes face challenges, including high costs, substantial time investments, potential cognitive biases, and limitations in exploring vast hypothesis spaces or complex data interactions systematically~\citep{kitano2021nobel}. Additionally, the increasing complexity and volume of data generated in modern research demand novel approaches to efficiently analyze and interpret such extensive raw information~\citep{miolane2025fifth, chitturi2024targeted}.

To mitigate these limitations, scientific agents have been developed to automate and accelerate critical aspects of scientific discovery~\citep{gridach2025agentic}. Such agents are designed to systematically handle experimentation, hypothesis formulation, data acquisition, analysis, and interpretation, thereby potentially enhancing research efficiency and reducing human-induced errors or biases. Computational agents promise greater scalability and consistency in scientific exploration, opening doors to novel discoveries that might otherwise remain obscured~\citep{szymanski2023autonomous,ghafarollahi2024protagents,liu2024toward, chen2023chemist, gottweis2025aicoscientist}. 

Early computational scientific agents were often designed for specific, narrowly focused tasks, lacking the adaptability required to generalize across varying research areas or to handle unforeseen challenges~\citep{beeler2024chemgymrl, maurizi2024inverse, rajak2021autonomous, dressler2018reinforcement, liang2020adaptive}. These systems were effective for well-defined problems but were not equipped to engage in the broader, more flexible processes of scientific exploration. The need for a more adaptive and powerful approach was clear.

Recent breakthroughs in in large language models (LLMs) offer potential for scientific discovery by introducing unprecedented capabilities in reasoning, planning, and handling complex, multimodal contexts. After being trained on vast amounts of textual data, LLMs provide a unified framework capable of understanding and orchestrating diverse languages, including natural language, computer language, and even physical information (e.g. symbolic representations of physical phenomena)~\citep{luo2022biogpt,dunn2024foundational,huang2024trustllm}. The integrative power of LLMs allows them to reason over scientific data in multiple modalities~\citep{mCLM2025}, seamlessly interact with humans, computational tools, and physical instruments alike, fostering more intuitive, flexible, and efficient workflows in scientific discovery. For example, \citet{Zhang2024Survey} systematically reviews more than 260 scientific LLMs across general science, mathematics, physics, chemistry, materials science, biology, medicine, and geoscience. It introduces how LLMs are being used to augment the scientific discovery process. \citet{reddy2025towards} argues that, while LLMs excel at specific scientific tasks, integrated systems capable of supporting the full autonomous scientific discovery cycle are still lacking. 

Fueled by advances in LLMs, research on scientific agents has grown rapidly, pointing toward a more autonomous paradigm for scientific discovery. This approach fully leverages the various capabilities of LLMs at every stage of the scientific discovery cycle. While several surveys provide broad overviews of general-purpose LLM agents~\citep{wang2024survey,guo2024large,mialon2023augmented}, only a few have focused specifically on their roles in scientific research~\citep{gao2024empowering,ren2025towards,wei2025ai,zhangcollective,zheng2025automation,luo2025llm4sr}. \citet{gao2024empowering} posits AI agents as the next frontier in scientific discovery, proposing a framework of autonomy levels based on their degree of participation in the scientific process. It introduces the core modules of agents that enable them to learn, plan, and engage in the scientific process. \citet{ren2025towards} discusses related work on scientific agents by focusing on the different implementation methods for each of the core modules of agents. \citet{wei2025ai} primarily focuses the correspondence between different stages of the scientific discovery cycle and the capabilities of LLM agents. It organizes relevant scientific discovery tasks and agent-based research by four major fields, including life sciences, chemistry, physics, and materials science. \citet{luo2025llm4sr} extends scientific research to include aspects such as paper writing and peer review. \citet{zheng2025automation} proposes a similar framework of agent levels and stages of scientific discovery to previous works \citet{gao2024empowering,wei2025ai}. Moving beyond individual agents, \citet{zhangcollective} argues that multi-agent systems can achieve a collective intelligence that surpasses the capabilities of a single agent by emulating the collaborative dynamics of human research teams. The work also details the advantages, limitations, and future directions of multi-agent systems in multiple stages of scientific discovery.

\begin{figure*}[t] 
    \centering 
    \includegraphics[width=\textwidth]{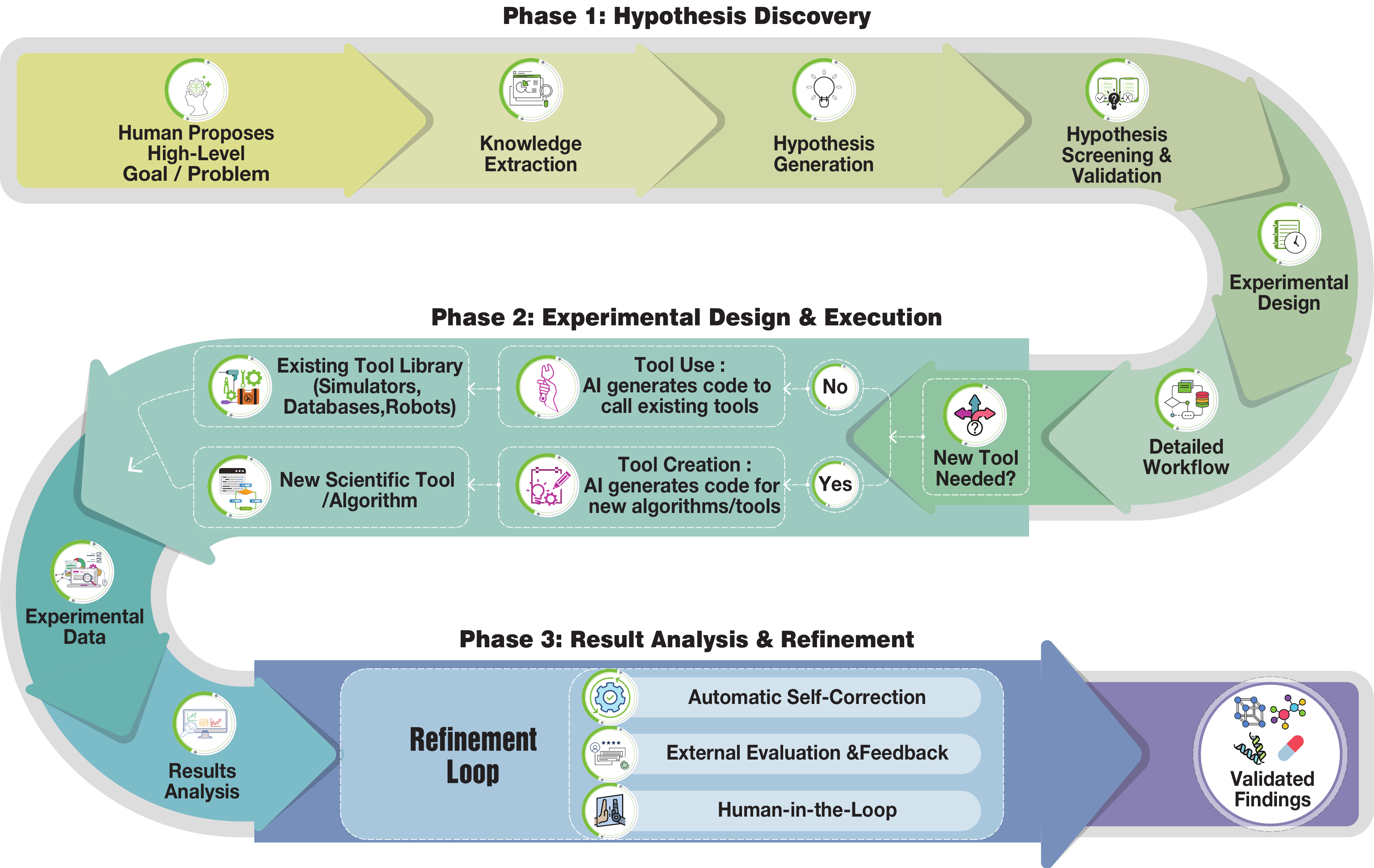} 
    \caption{An overview of the three-phase workflow for AI-driven scientific discovery. The process begins with Phase 1: Hypothesis Discovery, where a high-level human goal is transformed through knowledge extraction and hypothesis generation into novel, verifiable scientific questions. In Phase 2: Experimental Design \& Execution, these hypotheses are translated into detailed workflows which the agent executes by either using existing tools or creating new scientific tools to generate experimental data. Finally, Phase 3: Result Analysis \& Refinement involves interpreting the data and entering an iterative refinement loop to progressively improve the process and arrive at validated findings.}
    \label{fig:overview} 
\end{figure*}

Here, our work systematically examines how LLM-based agents operate across key stages of scientific discovery, including \textbf{hypothesis discovery}, \textbf{experimental design and execution}, and \textbf{result analysis and refinement}. Figure\ref{fig:overview} provides an overview of the three-phase workflow for AI-driven scientific discovery. We highlight the unique strengths of LLMs in enabling more adaptive, flexible, and capable scientific agents compared to traditional computational approaches. Furthermore, we critically assess existing methodologies, underscoring both their practical achievements and prevailing limitations. We develop a taxonomy of the methods used by scientific agents in each phase of scientific discovery and categorize the applications and achievements of scientific agents by domains (Figure \ref{fig:Related Work}). Finally, we identify open challenges and outline future research directions to advance the development of robust, generalizable, and highly adaptive LLM-driven scientific agents.

\begin{figure}[htp]
    \centering 
    \resizebox{0.95\textwidth}{!}{%
        \begin{tikzpicture}[node distance=2mm]

            \node[anchor=west, font=\footnotesize] (hd_main_label) {\textbf{Hypothesis Discovery}};
            
            \coordinate (start_pos) at ($(hd_main_label.south west) + (0.15\textwidth + 2mm + 0.2\textwidth + 2mm, -1mm)$);

            \node[egbox=lightblue, anchor=north west] (hd_ke_sfm_eg) at (start_pos) {\eglist{BioBERT~\citep{lee2020biobert}, BioGPT~\citep{luo2022biogpt}, Galactica~\citep{taylor2022galactica}, ether0~\citep{narayanan2025training},ChemDFM ~\citep{zhao2025developing},LPMs~\citep{barman2025large}}};
            \node[egbox=lightblue, anchor=north west, below=of hd_ke_sfm_eg] (hd_ke_rag_eg) {\eglist{CLADD~\citep{lee2025rag}, CASSIA~\citep{xie2024cassia},FRAG~\citep{gao2025frag},CLEAR~\citep{lopez2025clinical}}};
            \node[egbox=lightblue, anchor=north west, below=of hd_ke_rag_eg] (hd_ke_mke_eg) {\eglist{ChemMiner~\citep{chen2024autonomous}, ChartAssistant~\citep{meng2024chartassisstant}, GPT-4o~\citep{hurst2024gpt},MatViX~\citep{khalighinejad2024matvix},nanoMINER~\cite{odobesku2025agent}}};

            \path let \p1=(hd_ke_sfm_eg.north), \p2=(hd_ke_sfm_eg.south), \n1={\y1-\y2} in
                node[methodbox=lightblue, minimum height=\n1, anchor=north east, at=(hd_ke_sfm_eg.north west), xshift=-2mm] (hd_ke_sfm) {Scientific Foundation Models};
            \path let \p1=(hd_ke_rag_eg.north), \p2=(hd_ke_rag_eg.south), \n1={\y1-\y2} in
                node[methodbox=lightblue, minimum height=\n1, anchor=north east, at=(hd_ke_rag_eg.north west), xshift=-2mm] (hd_ke_rag) {RAG-based};
            \path let \p1=(hd_ke_mke_eg.north), \p2=(hd_ke_mke_eg.south), \n1={\y1-\y2} in
                node[methodbox=lightblue, minimum height=\n1, anchor=north east, at=(hd_ke_mke_eg.north west), xshift=-2mm] (hd_ke_mke) {Multimodal Knowledge Extraction};

            \path let \p1=(hd_ke_sfm_eg.north), \p2=(hd_ke_mke_eg.south), \n1={\y1-\y2} in
                node[subbox=lightblue, minimum height=\n1, anchor=north east, at=(hd_ke_sfm.north west), xshift=-2mm] (hd_ke) {Knowledge Extraction};

            \node[egbox=lightblue, anchor=north west] (hd_hg_pbs_eg) at ($(hd_ke_mke_eg.south west) + (0, -1mm)$) {\eglist{Biomedicine~\citep{qi2023large}, General Science~\citep{xiong2025toward}}};
            \node[egbox=lightblue, anchor=north west, below=of hd_hg_pbs_eg] (hd_hg_kg_eg) {\eglist{KG-CoI~\citep{xiong2024improving}, ResearchLink~\citep{borrego2025research}}};
            \node[egbox=lightblue, anchor=north west, below=of hd_hg_kg_eg] (hd_hg_mas_eg) {\eglist{AI co-scientist~\citep{gottweis2025aicoscientist}, ACCELMAT~\citep{kumbhar2025hypothesisgenerationmaterialsdiscovery}, VIRSCI~\citep{su2024many}, AstroAgents~\citep{saeedi2025astroagents}}};
            \node[egbox=lightblue, anchor=north west, below=of hd_hg_mas_eg] (hd_hg_ea_eg) {\eglist{MOOSE-Chem~\citep{yang2025moosechemlargelanguagemodels}, HypoAgents~\citep{duan2025bayes}, MolLEO \citep{wang2024efficient}}};

            \path let \p1=(hd_hg_pbs_eg.north), \p2=(hd_hg_pbs_eg.south), \n1={\y1-\y2} in
                node[methodbox=lightblue, minimum height=\n1, anchor=north east, at=(hd_hg_pbs_eg.north west), xshift=-2mm] (hd_hg_pb) {Prompt-Based};
            \path let \p1=(hd_hg_kg_eg.north), \p2=(hd_hg_kg_eg.south), \n1={\y1-\y2} in
                node[methodbox=lightblue, minimum height=\n1, anchor=north east, at=(hd_hg_kg_eg.north west), xshift=-2mm] (hd_hg_kg) {Knowledge-Grounded};
            \path let \p1=(hd_hg_mas_eg.north), \p2=(hd_hg_mas_eg.south), \n1={\y1-\y2} in
                node[methodbox=lightblue, minimum height=\n1, anchor=north east, at=(hd_hg_mas_eg.north west), xshift=-2mm] (hd_hg_mas) {Multi-Agent};
            \path let \p1=(hd_hg_ea_eg.north), \p2=(hd_hg_ea_eg.south), \n1={\y1-\y2} in
                node[methodbox=lightblue, minimum height=\n1, anchor=north east, at=(hd_hg_ea_eg.north west), xshift=-2mm] (hd_hg_ea) {Evolutionary Algorithm-Based};

            \path let \p1=(hd_hg_pbs_eg.north), \p2=(hd_hg_ea_eg.south), \n1={\y1-\y2} in
                node[subbox=lightblue, minimum height=\n1, anchor=north east, at=(hd_hg_pb.north west), xshift=-2mm] (hd_hg) {Hypothesis Generation};

            \node[egbox=lightblue, anchor=north west] (hd_hsv_mbs_eg) at ($(hd_hg_ea_eg.south west) + (0, -2mm)$) {\eglist{POPPER~\citep{huang2025automatedhypothesisvalidationagentic}, SCIMON~\citep{wang2024scimon}, MATDESIGN~\citep{kumbhar2025hypothesis}, HypoBench~\citep{liu2025hypobench}, Logit-based Calibrated Prior~\citep{gong2025exploiting}}};
            \node[egbox=lightblue, anchor=north west, below=of hd_hsv_mbs_eg] (hd_hsv_abv_eg) {\eglist{ResearchAgent~\citep{baek2024researchagent}, Scientific Generative Agent (SGA)~\citep{ma2024llm}}};

            \path let \p1=(hd_hsv_mbs_eg.north), \p2=(hd_hsv_mbs_eg.south), \n1={\y1-\y2} in
                node[methodbox=lightblue, minimum height=\n1, anchor=north east, at=(hd_hsv_mbs_eg.north west), xshift=-2mm] (hd_hsv_mbs) {Metric-Based Screening};
            \path let \p1=(hd_hsv_abv_eg.north), \p2=(hd_hsv_abv_eg.south), \n1={\y1-\y2} in
                node[methodbox=lightblue, minimum height=\n1, anchor=north east, at=(hd_hsv_abv_eg.north west), xshift=-2mm] (hd_hsv_abv) {Agent-Based Validation};

            \path let \p1=(hd_hsv_mbs_eg.north), \p2=(hd_hsv_abv_eg.south), \n1={\y1-\y2} in
                node[subbox=lightblue, minimum height=\n1, anchor=north east, at=(hd_hsv_mbs.north west), xshift=-2mm] (hd_hsv) {Hypothesis Screening \\ \& Validation};
            
            \begin{scope}[on background layer]
                \node[mainbox=mainblue, fit=(hd_main_label) (hd_hsv) (hd_hsv_abv_eg)] (block1_fit) {};
            \end{scope}

            \node[anchor=west, font=\footnotesize] (ede_main_label) at ($(hd_main_label.west |- block1_fit.south) + (0,-0.4cm)$) {\textbf{Experimental Design \& Execution}};
            
            \coordinate (block2_start_pos) at (start_pos |- ede_main_label.south) ++(0,-1mm);

            \node[egbox=mainred, anchor=north west] (ede_ed_rag_eg) at (block2_start_pos) {\eglist{Biomni~\citep{huang2025biomni}, TxAgent~\citep{gao2025txagent}, Chemist-X~\citep{chen2023chemist}, CLADD~\citep{lee2025rag}}};
            \node[egbox=mainred, anchor=north west, below=of ede_ed_rag_eg] (ede_ed_hhg_eg) {\eglist{CRISPR-GPT~\citep{huang2025crispr}, AI co-scientist~\citep{gottweis2025aicoscientist}, Virtual Lab~\citep{swanson2024virtual}}};
            \node[egbox=mainred, anchor=north west, below=of ede_ed_hhg_eg] (ede_ed_tpa_eg) {\eglist{k-agents~\citep{cao2024agents}, CRISPR-GPT~\citep{huang2025crispr}, PerTurboAgent~\citep{hao2025perturboagent}}};
            \node[egbox=mainred, anchor=north west, below=of ede_ed_tpa_eg] (ede_ed_pef_eg) {\eglist{ChemToolAgent~\citep{yu2024chemtoolagent}, BioInformatics Agent~\citep{xin2024bioinformatics}, BioDiscoveryAgent~\citep{roohani2024biodiscoveryagent}, OSDA Agent~\citep{hu2025osda}}};

            \path let \p1=(ede_ed_rag_eg.north), \p2=(ede_ed_rag_eg.south), \n1={\y1-\y2} in
                node[methodbox=mainred, minimum height=\n1, anchor=north east, at=(ede_ed_rag_eg.north west), xshift=-2mm] (ede_ed_rag) {RAG-based Planning};
            \path let \p1=(ede_ed_hhg_eg.north), \p2=(ede_ed_hhg_eg.south), \n1={\y1-\y2} in
                node[methodbox=mainred, minimum height=\n1, anchor=north east, at=(ede_ed_hhg_eg.north west), xshift=-2mm] (ede_ed_hhg) {Human High-Level Guidance};
            \path let \p1=(ede_ed_tpa_eg.north), \p2=(ede_ed_tpa_eg.south), \n1={\y1-\y2} in
                node[methodbox=mainred, minimum height=\n1, anchor=north east, at=(ede_ed_tpa_eg.north west), xshift=-2mm] (ede_ed_tpa) {Templates \& Predefined Actions};
            \path let \p1=(ede_ed_pef_eg.north), \p2=(ede_ed_pef_eg.south), \n1={\y1-\y2} in
                node[methodbox=mainred, minimum height=\n1, anchor=north east, at=(ede_ed_pef_eg.north west), xshift=-2mm] (ede_ed_pef) {Post-Execution Feedback};

            \path let \p1=(ede_ed_rag_eg.north), \p2=(ede_ed_pef_eg.south), \n1={\y1-\y2} in
                node[subbox=mainred!60, minimum height=\n1, anchor=north east, at=(ede_ed_rag.north west), xshift=-2mm] (ede_ed) {Experimental Design};

            \node[egbox=mainred, anchor=north west] (em_eg) at ($(ede_ed_pef_eg.south west) + (0, -1mm)$) {\eglist{OSDA Agent~\citep{hu2025osda}, LLMatDesign~\citep{jia2024llmatdesign}, MatLLMSearch~\citep{gan2025large}, scBaseCount~\citep{youngblut2025scbasecount}, Chemist-X~\citep{chen2023chemist}}};
            \node[egbox=mainred, anchor=north west, below=of em_eg] (tb_eg) {\eglist{AgentMD~\citep{jin2024agentmd}, ChemCrow~\citep{m2024augmenting}, scAgent~\citep{mao2025scagent}, TxAgent~\citep{gao2025txagent}, GIS Copilot~\citep{akinboyewa2025gis}, SciToolAgent~\citep{chen2025scitoolagent}}};
            \node[egbox=mainred, anchor=north west, below=of tb_eg] (ri_eg) {\eglist{ChemToolAgent~\citep{yu2024tooling}, Biomni~\citep{huang2025biomni}, PerTurboAgent~\citep{hao2025perturboagent}}};
            \node[egbox=mainred, anchor=north west, below=of ri_eg] (mas_eg) {\eglist{TAIS~\citep{liu2024toward}, MedAgents~\citep{tang2023medagents}, ChemAgents~\citep{song2025multiagent}, AtomAgent~\citep{ghafarollahi2024atomagentsalloydesigndiscovery}, ProtAgents~\citep{ghafarollahi2024protagents}, ChatGPT Research Group~\citep{zheng2023chatgpt}, Virtual Lab~\citep{swanson2024virtual}, k-agents~\citep{cao2024agents}}};

            \path let \p1=(em_eg.north), \p2=(em_eg.south), \n1={\y1-\y2} in
                node[methodbox=mainred, minimum height=\n1, anchor=north east, at=(em_eg.north west), xshift=-2mm] (em) {Embedded};
            \path let \p1=(tb_eg.north), \p2=(tb_eg.south), \n1={\y1-\y2} in
                node[methodbox=mainred, minimum height=\n1, anchor=north east, at=(tb_eg.north west), xshift=-2mm] (tb) {Toolbox-Based};
            \path let \p1=(ri_eg.north), \p2=(ri_eg.south), \n1={\y1-\y2} in
                node[methodbox=mainred, minimum height=\n1, anchor=north east, at=(ri_eg.north west), xshift=-2mm] (ri) {Reflective \& Iterative};
            \path let \p1=(mas_eg.north), \p2=(mas_eg.south), \n1={\y1-\y2} in
                node[methodbox=mainred, minimum height=\n1, anchor=north east, at=(mas_eg.north west), xshift=-2mm] (mas) {Hierarchical in MAS};

            \path let \p1=(em_eg.north), \p2=(mas_eg.south), \n1={\y1-\y2} in
                node[subbox=mainred!60, minimum height=\n1, anchor=north east, at=(em.north west), xshift=-2mm] (tu) {Tool Use};

            \node[egbox=mainred, anchor=north west, text width=\dimexpr 0.2\textwidth + 0.56\textwidth + 3mm\relax] (tc_eg) at ($(mas.west |- mas_eg.south) + (0,-1mm)$) {\eglist{ToolUniverse~\citep{gao2025democratizing}, MAPPS~\citep{zhou2025greaterautonomymaterialsdiscovery}, CodePDE~\citep{li2025codepdeinferenceframeworkllmdriven}, AlphaEvolve~\citep{novikov2025alphaevolvecodingagentscientific}, TOOLMAKER~\citep{wolflein2025llm}, ShinkaEvolve~\cite{lange2025shinkaevolve}}};
            \path let \p1=(tc_eg.north), \p2=(tc_eg.south), \n1={\y1-\y2} in
                node[subbox=mainred!60, minimum height=\n1, anchor=north east, at=(tc_eg.north west), xshift=-2mm] (tc) {Tool Creation};
            
            \begin{scope}[on background layer]
                \node[mainbox=mainred, fit=(ede_main_label) (tc) (tc_eg)] (block2_fit) {};
            \end{scope}

            \node[anchor=west, font=\footnotesize] (rar_main_label) at ($(hd_main_label.west |- block2_fit.south) + (0,-0.4cm)$) {\textbf{Result Analysis \& Refinement}};
            
            \coordinate (block3_start_pos) at (start_pos |- rar_main_label.south) ++(0,-1mm);

            \node[egbox=lightblue, anchor=north west] (rar_ra_mda_eg) at (block3_start_pos) {\eglist{The AI Scientist-v2~\citep{yamada2025ai}, PlotGen~\citep{luo2024plotgen}, ChartLlama~\citep{han2023chartllama}, ChartAssisstant~\citep{meng2024chartassisstant}}};
            \node[egbox=lightblue, anchor=north west, below=of rar_ra_mda_eg] (rar_ra_tea_eg) {\eglist{ChemCrow~\citep{bran2023chemcrow}, Coscientist~\citep{boiko2023autonomous}, Virtual Lab~\citep{boeck2023multiagent}}};
            \node[egbox=lightblue, anchor=north west, below=of rar_ra_tea_eg] (rar_ra_cna_eg) {\eglist{Data Interpreter~\citep{hong2024data}, LLM-SR~\citep{kamienny2024llmsr}, MOBLLM~\citep{binbas2024autonomous}}};

            \path let \p1=(rar_ra_mda_eg.north), \p2=(rar_ra_mda_eg.south), \n1={\y1-\y2} in
                node[methodbox=lightblue, minimum height=\n1, anchor=north east, at=(rar_ra_mda_eg.north west), xshift=-2mm] (rar_ra_mda) {Modality-Driven};
            \path let \p1=(rar_ra_tea_eg.north), \p2=(rar_ra_tea_eg.south), \n1={\y1-\y2} in
                node[methodbox=lightblue, minimum height=\n1, anchor=north east, at=(rar_ra_tea_eg.north west), xshift=-2mm] (rar_ra_tea) {Tool-Augmented};
            \path let \p1=(rar_ra_cna_eg.north), \p2=(rar_ra_cna_eg.south), \n1={\y1-\y2} in
                node[methodbox=lightblue, minimum height=\n1, anchor=north east, at=(rar_ra_cna_eg.north west), xshift=-2mm] (rar_ra_cna) {Computation-Native};

            \path let \p1=(rar_ra_mda_eg.north), \p2=(rar_ra_cna_eg.south), \n1={\y1-\y2} in
                node[subbox=lightblue, minimum height=\n1, anchor=north east, at=(rar_ra_mda.north west), xshift=-2mm] (rar_ra) {Result Analysis};

            \node[egbox=lightblue, anchor=north west] (rar_ivr_asc_eg) at ($(rar_ra_cna_eg.south west) + (0, -1mm)$) {\eglist{Self-Refine~\citep{madaan2023self}, ChemAgent~\citep{tang2025chemagentselfupdatinglibrarylarge}, SpatialAgent~\citep{wang2025spatialagent}, TAIS~\citep{liu2024toward}, BioInformatics Agent~\citep{xin2024bioinformatics}}};
            \node[egbox=lightblue, anchor=north west, below=of rar_ivr_asc_eg] (rar_ivr_eef_eg) {\eglist{CRITIC~\citep{gou2023critic}, OSDA Agent~\citep{hu2025osda}, AI co-scientist~\citep{gottweis2025aicoscientist}, BioDiscoveryAgent~\citep{roohani2024biodiscoveryagent}, SAMPLE~\citep{rapp2024self}, Biomni~\citep{huang2025biomni}}};
            \node[egbox=lightblue, anchor=north west, below=of rar_ivr_eef_eg] (rar_ivr_hil_eg) {\eglist{AI co-scientist~\citep{gottweis2025aicoscientist}, ResearchAgent~\citep{baek2024researchagent}, MAPPS~\citep{zhou2025greaterautonomymaterialsdiscovery}}};

            \path let \p1=(rar_ivr_asc_eg.north), \p2=(rar_ivr_asc_eg.south), \n1={\y1-\y2} in
                node[methodbox=lightblue, minimum height=\n1, anchor=north east, at=(rar_ivr_asc_eg.north west), xshift=-2mm] (rar_ivr_asc) {Automatic Self-correction};
            \path let \p1=(rar_ivr_eef_eg.north), \p2=(rar_ivr_eef_eg.south), \n1={\y1-\y2} in
                node[methodbox=lightblue, minimum height=\n1, anchor=north east, at=(rar_ivr_eef_eg.north west), xshift=-2mm] (rar_ivr_eef) {External Evaluation \& Feedback};
            \path let \p1=(rar_ivr_hil_eg.north), \p2=(rar_ivr_hil_eg.south), \n1={\y1-\y2} in
                node[methodbox=lightblue, minimum height=\n1, anchor=north east, at=(rar_ivr_hil_eg.north west), xshift=-2mm] (rar_ivr_hil) {Human-in-the-Loop};

            \path let \p1=(rar_ivr_asc_eg.north), \p2=(rar_ivr_hil_eg.south), \n1={\y1-\y2} in
                node[subbox=lightblue, minimum height=\n1, anchor=north east, at=(rar_ivr_asc.north west), xshift=-2mm] (rar_ivr) {Iterative \\ Validation \\\& Refinement};
            
            \begin{scope}[on background layer]
                \node[mainbox=mainblue, fit=(rar_main_label) (rar_ivr) (rar_ivr_hil_eg)] (block3_fit) {};
            \end{scope}

        \end{tikzpicture}%
    } 
    
\end{figure}

\begin{figure}[h!] 
    \centering 
    \resizebox{0.96\textwidth}{!}{%
        \begin{tikzpicture}[node distance=2mm]
            \node[anchor=west, font=\footnotesize] (dsa_main_label) {\textbf{Domain-Specific Scientific Agents}};
            \coordinate (dsa_start_pos) at ($(dsa_main_label.south west) + (0.15\textwidth + 2mm, -1mm)$);

            \node[egbox=mainpurple, anchor=north west, text width=\dimexpr 0.81\textwidth\relax] (dsa_gen_eg) at (dsa_start_pos) {\eglist{SpatialAgent~\citep{wang2025spatialagent}, TAIS~\citep{liu2024toward}, CRISPR-GPT~\citep{huang2025crispr}, BioDiscoveryAgent~\citep{roohani2024biodiscoveryagent}, PerTurboAgent~\citep{hao2025perturboagent}, BioAgents~\citep{mehandru2025bioagents}, BIA~\citep{xin2024bioinformatics}, scBaseCount~\citep{youngblut2025scbasecount}, CellAgent~\citep{xiao2024cellagent}, CompBioAgent~\citep{zhang2025compbioagent}, scAgent~\citep{mao2025scagent}, CASSIA~\citep{xie2024cassia}}};
            
            \node[egbox=mainpurple, anchor=north west, text width=\dimexpr 0.81\textwidth\relax, below=of dsa_gen_eg] (dsa_pro_eg) {\eglist{SAMPLE~\citep{rapp2024self}, Virtual Lab~\citep{swanson2024virtual}, ProtAgents~\citep{ghafarollahi2024protagents}, Sparks~\citep{ghafarollahi2025sparks}, VibeGen~\citep{ni2025agentic}, AutoProteinEngine (AutoPE)~\citep{liu2024autoproteinengine}, ProChat~\citep{huang2024protchat}}};
            
            \node[egbox=mainpurple, anchor=north west, text width=\dimexpr 0.81\textwidth\relax, below=of dsa_pro_eg] (dsa_bio_eg) {\eglist{Biomni~\citep{huang2025biomni}, AI co-scientist~\citep{gottweis2025towards}, TxAgent~\citep{gao2025txagent}, BioResearcher~\citep{luo2025intention}, STELLA~\citep{jin2025stella}, CLADD~\citep{lee2025rag}, DrugAgent~\citep{liu2024drugagent}, PharmAgents~\citep{gao2025pharmagents}, LIDDiA~\citep{averly2025liddia}, AgentMD~\citep{jin2024agentmd}, MedAgents~\citep{tang2023medagents}, ClinicalGPT~\citep{wang2023clinicalgpt}, BehaveAgent~\citep{aljovic2025autonomous}}};
            
            \node[egbox=mainpurple, anchor=north west, text width=\dimexpr 0.81\textwidth\relax, below=of dsa_bio_eg] (dsa_che_eg) {\eglist{Coscientist~\citep{boiko2023autonomous}, ChemCrow~\citep{m2024augmenting}, ChemAgents~\citep{song2025multiagent}, ChemReasoner~\citep{sprueill2024chemreasoner}, CACTUS~\citep{mcnaughton2024cactus}, Chemist-X~\citep{chen2023chemist}, El Agente Q~\citep{zou2025agente}, LLM-RDF~\citep{ruan2024automatic}, FROGENT~\citep{pan2025frogent}, LARC~\citep{baker2025larc}, FMG \citep{sun2025foundation}, MOOSE-Chem~\citep{yang2024moose}}};
            
            \node[egbox=mainpurple, anchor=north west, text width=\dimexpr 0.81\textwidth\relax, below=of dsa_che_eg] (dsa_mat_eg) {\eglist{A-Lab~\citep{szymanski2023autonomous}, ChatMOF~\citep{kang2024chatmof}, ChatGPT Research Group~\citep{zheng2023chatgpt}, MDAgent~\citep{shi2025fine}, HoneyComb~\citep{zhang2024honeycomb}, LLMatDesign~\citep{jia2024llmatdesign}, MatAgent~\citep{bazgir2025matagent}, MatSciAgent~\citep{chaudhari2025modular}, MatPilot~\citep{ni2024matpilot}, Materials Laws Multi-Agent Framework~\citep{hu2024multi}}};
            
            \node[egbox=mainpurple, anchor=north west, text width=\dimexpr 0.81\textwidth\relax, below=of dsa_mat_eg] (dsa_phy_eg) {\eglist{k-agents~\citep{cao2024agents}, AtomAgents~\citep{ghafarollahi2024atomagents}, Mephisto~\citep{sun2024interpreting}, QCopilot~\citep{sha2025llm}, OpenFOAMGPT~\citep{pandey2025openfoamgpt}, MetaOpenFOAM~\citep{chen2024metaopenfoam}, AI-Scientist Framework~\citep{xu2025advancing}}};

            \node[egbox=mainpurple, anchor=north west, text width=\dimexpr 0.81\textwidth\relax, below=of dsa_phy_eg] (dsa_eng_eg) {\eglist{Autonomous GIS Agent~\citep{ning2025autonomous}, GIS Copilot~\citep{akinboyewa2025gis}, LP-COMDA~\citep{liu2024physics}, ReAct agent for gas turbines~\citep{song2024domain}}};           
            
            \path let \p1=(dsa_gen_eg.north), \p2=(dsa_gen_eg.south), \n1={\y1-\y2} in
                node[subbox=mainpurple!60, minimum height=\n1, anchor=north east, at=(dsa_gen_eg.north west), xshift=-2mm] (dsa_gen) {Genomics};
            \path let \p1=(dsa_pro_eg.north), \p2=(dsa_pro_eg.south), \n1={\y1-\y2} in
                node[subbox=mainpurple!60, minimum height=\n1, anchor=north east, at=(dsa_pro_eg.north west), xshift=-2mm] (dsa_pro) {Protein};
            \path let \p1=(dsa_bio_eg.north), \p2=(dsa_bio_eg.south), \n1={\y1-\y2} in
                node[subbox=mainpurple!60, minimum height=\n1, anchor=north east, at=(dsa_bio_eg.north west), xshift=-2mm] (dsa_bio) {Medicine};
            \path let \p1=(dsa_che_eg.north), \p2=(dsa_che_eg.south), \n1={\y1-\y2} in
                node[subbox=mainpurple!60, minimum height=\n1, anchor=north east, at=(dsa_che_eg.north west), xshift=-2mm] (dsa_che) {Chemistry};
            \path let \p1=(dsa_mat_eg.north), \p2=(dsa_mat_eg.south), \n1={\y1-\y2} in
                node[subbox=mainpurple!60, minimum height=\n1, anchor=north east, at=(dsa_mat_eg.north west), xshift=-2mm] (dsa_mat) {Materials};
            \path let \p1=(dsa_phy_eg.north), \p2=(dsa_phy_eg.south), \n1={\y1-\y2} in
                node[subbox=mainpurple!60, minimum height=\n1, anchor=north east, at=(dsa_phy_eg.north west), xshift=-2mm] (dsa_phy) {Physics};
            \path let \p1=(dsa_eng_eg.north), \p2=(dsa_eng_eg.south), \n1={\y1-\y2} in
                node[subbox=mainpurple!60, minimum height=\n1, anchor=north east, at=(dsa_eng_eg.north west), xshift=-2mm] (dsa_eng) {Others};
                
            \begin{scope}[on background layer]
                \node[mainbox=mainpurple, fit=(dsa_main_label) (dsa_eng) (dsa_eng_eg)] (block4_fit) {};
            \end{scope}
        \end{tikzpicture}%
    }
    \caption{A comprehensive overview of the role of LLM-based agents in the scientific discovery lifecycle alongside a diverse collection of domain-specific scientific agents  that organize various research systems and papers across fields such as Genomics, Protein, Medicine, Chemistry, Materials, Physics, and Others.}
\label{fig:Related Work}
\vspace{-1em}
\end{figure}

\section{Overview of Autonomous Agents for Scientific Discovery}

\subsection{Overview of Scientific Discovery}
To understand how LLM agents transform scientific discovery, we first revisit how scientific processes are typically structured across domains. 
Scientific discovery is the systematic pursuit of understanding and generating new knowledge or verifying hypotheses within scientific domains. It involves identifying meaningful questions, formulating hypotheses~\citep{scimon2024}, generating experiment procedures~\citep{actionie2024}, conducting experiments, and refining results iteratively to enhance understanding or achieve practical outcomes. While objectives differ across fields, such as identifying stable crystal structures in materials science, elucidating genetic mechanisms in biology, or systematically synthesizing therapeutic compounds in pharmacology, the discovery process shares common core components and phases across disciplines. We conceptualize the scientific discovery processes into three phases.

\textbf{Hypothesis Discovery.}
This initial phase is a crucial creative process with the core objective of identifying and forming novel, verifiable scientific hypotheses from vast amounts of data and existing knowledge. It aims to accelerate the pace of discovery by revealing hidden connections that are difficult for human researchers to find on their own. A complete hypothesis discovery workflow is typically composed of three key stages, including Knowledge Extraction, Hypothesis Generation, and Hypothesis Screening and Validation. The goal is to systematically identify meaningful questions and formulate hypotheses that can be tested in subsequent stages.

\textbf{Experimental Design and Execution.}
Following hypothesis discovery, experimental design is a critical step that involves creating a structured plan to systematically test ideas and achieve research objectives. This process is a form of workflow generation where high-level scientific goals are translated into concrete, executable protocols. The execution stage serves as the crucial bridge that transforms abstract strategies into concrete actions and empirical evidence needed for generating insights. It involves complex tasks such as orchestrating computational resources, managing large datasets, interfacing with laboratory instruments or simulators, and continuously monitoring the workflow. It's important to clarify that our definition of experiments includes both traditional wet-lab experiments and computational simulations. While both are integral to the scientific process, current LLM-based scientific agents predominantly operate in the computational domain, focusing on simulation-based discovery. LLM-based agents accomplish this primarily through two interconnected capabilities, namely tool use and tool creation.

\textbf{Result Analysis and Refinement.}
This phase begins with the analysis of experimental results, where autonomous agents interpret raw outputs to derive meaningful scientific insights. Since scientific discovery rarely concludes after a single iteration, this stage functions as a crucial iterative mechanism. It involves cycles of reviewing validated results, identifying discrepancies or unexpected findings, and then methodically refining the proposed hypotheses, experimental design, or computational tools to progressively enhance scientific outcomes.

Collectively, these stages encapsulate the iterative and cyclical nature of scientific inquiry, continuously refining and verifying insights to produce credible and innovative scientific knowledge.

\subsection{Overview of Scientific Agents}
A scientific agent is a specialized AI system designed to simulate and autonomously execute aspects of the scientific research process. Unlike general-purpose LLM agents \citep{chatgpt,gemini,claude}, which are built for a wide array of tasks like conversation and text generation, scientific agents are highly domain-specific and task-oriented. Their primary goal is to accelerate scientific discovery by automating and augmenting the capabilities of human researchers, thereby allowing the navigation of the intricate landscape of modern science with greater speed and efficiency. Table~\ref{tab:agent_comparison} summarizes the key differences between scientific agents and general-purpose LLM agents.

\begin{table}[t!]
\centering
\caption{Comparison of Scientific AI Agents and General LLM Agents}
\begin{tblr}{
  colspec={Q[l,m,0.1\linewidth]Q[l,m,0.35\linewidth]Q[l,m,0.35\linewidth]},
  column{1}={bg=lightgray},
  column{2}={bg=lightblue},
  column{3}={bg=lightgreen},
  hlines,
  vlines={},
  cells={valign=m, font=\small},
}

\textbf{Feature} & \textbf{Scientific Agent} & \textbf{General LLM Agent} \\
Primary Objective & To solve specific, complex scientific problems and generate new knowledge. & To understand and respond to a wide range of general user queries and commands. \\
Knowledge Base & Deeply integrated with specialized scientific databases, literature, and models. & Relies on broad, general knowledge acquired from vast, diverse training data. \\
Reasoning & Employs rigorous, multi-step logical deduction and inference based on scientific principles. & Utilizes common sense, associative, and heuristic reasoning for broad tasks. \\
Tool Use & Natively uses highly specialized software (e.g. simulators) and hardware (e.g. robots). & Interacts with general-purpose tools like web search engines and code interpreters. \\
Memory & Utilizes memory to accumulate domain knowledge and learn from experimental outcomes. & Memory is primarily used for maintaining conversational context and user preferences. \\
Evaluation Metrics & Success is measured by the accuracy, reproducibility, and novelty of the scientific results. & Success is measured by task completion, user satisfaction, and conversational quality. \\
\end{tblr}
\label{tab:agent_comparison}
\end{table}

\textbf{Reasoning and Planning.} The core cognitive functions of an LLM-based scientific agent are its advanced reasoning and planning abilities, which serve as its intellectual engine. Its reasoning capabilities allow it to sift through vast amount of scientific literature and data to formulate novel hypotheses, identify patterns, and draw logical inferences \citep{ferrag2025llm,zhang2024llm}. This is coupled with a sophisticated planning module that can decompose a high-level research goal into a coherent sequence of actionable steps \citep{wei2025plangenllms,huang2024understanding}.

\textbf{Tool Use.} A defining characteristic of LLM-based scientific agents is their deep integration with a wide array of specialized tools \citep{jin2024llms, shen2024llm}. This capability serves as the bridge between the agent's cognitive processes and the practical execution of research. These tools are not limited to software, such as domain-specific simulators, data analysis packages, and proprietary knowledge bases, but also extend to physical hardware. Through programmatic interfaces, scientific agents can operate laboratory automation systems, control sensor arrays, and manage other instrumentation, allowing them to directly interact with and conduct experiments in the physical world~\citep{boiko2023autonomous}. 

\textbf{Memory.} An LLM-based scientific agent's memory mechanism is fundamental to its capacity for sustained research and cumulative learning \citep{xu2025mem,wu2025human}. Acting as both a working memory for immediate tasks and a persistent knowledge base, it archives all past outcomes. By retaining successful protocols and failed inquiries, this long-term memory enables the agent to learn from experience. This ability to continuously reference its own history allows the agent to refine strategies, avoid repeating errors, and systematically drive the process of scientific discovery forward.

\subsection{An Information-Theoretic Framework for Autonomous Scientific Discovery}

\begin{figure*}[t]
    \centering
    \includegraphics[width=0.6\textwidth]{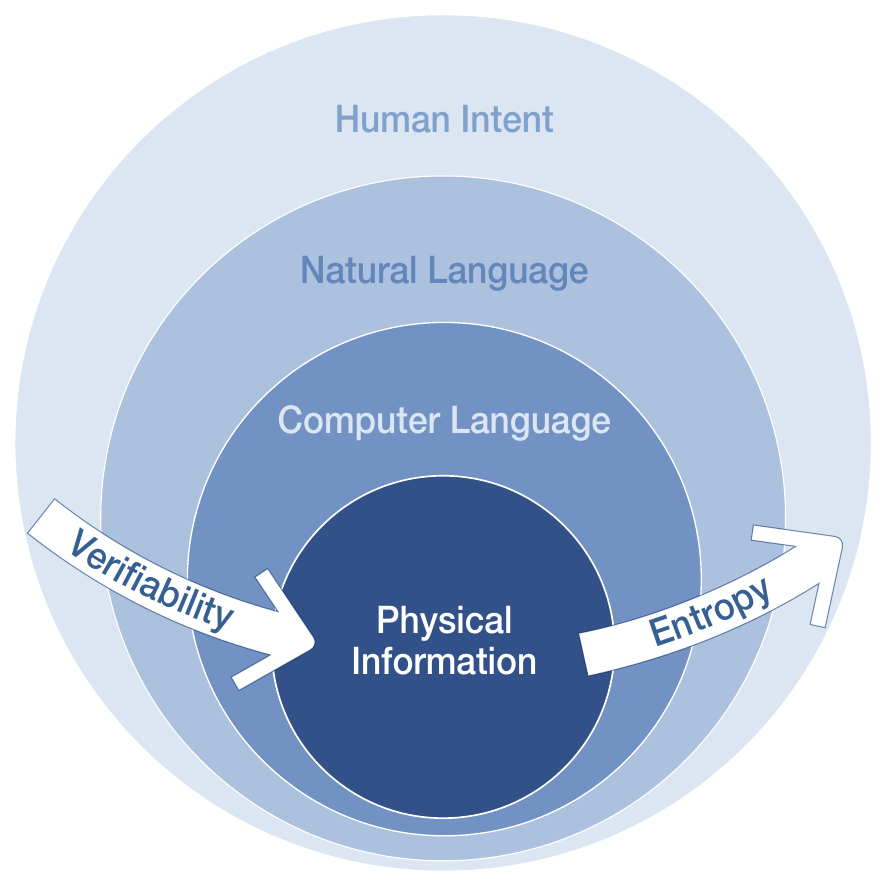}
    \caption{An information-theoretic framework for autonomous scientific discovery, illustrating the inverse relationship between Information Entropy and Verifiability.}
    \label{fig:InformationLevel}
\end{figure*}

\subsubsection{\label{sec:Information} Properties of Information in Autonomous Scientific Discovery}
The transition from human-led research to autonomous scientific discovery can be understood through a formal analytical framework. The core of this framework is to examine three key aspects of the information processed by an intelligent agent in the scientific workflow. First, we view scientific discovery as an inherently dissipative process, characterized by uncertainty and the need for extensive trial-and-error exploration. Second, we use the mathematical measure of Information Entropy to quantify the root of this uncertainty. Finally, we examine Verifiability, which measures whether information can be tested against objective standards and is the ultimate goal of the research process.

\textbf{Information Entropy} is a mathematical measure from information theory that quantifies the uncertainty or the size of the hypothesis space of a given problem. A high-entropy state, such as the set of all possible scientific hypotheses, presents a vast and unstructured search space for an agent. A core task for the agent is therefore to perform Entropy Reduction—to systematically reduce the uncertainty of the problem through its operations~\citep{yao2003information}.
Our notion of reducing information entropy is consistent with the Second Law of Thermodynamics, which states that the total entropy of an isolated system can never decrease~\citep{Borgnakke2024Fundamentals11,schroeder2020introduction}. For an agent to achieve Information Entropy Reduction in scientific discovery, the system must be an open system, instead of an isolated one. This systematic decrease in internal informational uncertainty is driven by the necessary exchange of information with the physical world. The agent's internal entropy reduction is fundamentally dependent on interactions with the physical world. The information required to constrain the hypothesis space cannot be generated internally. Instead, it must be obtained from real-world processes (\emph{e.g.}, performing experiments or analyzing raw data that originates from physical processes) to constrain uncertainty. 


\textbf{Verifiability} is a property of a specific information object that measures its ability to be objectively tested against a formal, logical, or empirical standard~\citep{patterson1978verifiability}. The ultimate goal of the scientific process is to transform an initial idea with low verifiability into an empirical fact with very high verifiability. The agent's workflow, therefore, is also a process of progressively imparting verifiability to information.

\textbf{Dissipation} refers to the unavoidable computational cost and effort expended on unproductive paths during the exploration of a problem space. This cost encompasses the resources consumed in the formulation, execution, and verification of incorrect hypotheses and experiments that ultimately do not lead to a solution.
In thermodynamics, a dissipative process is defined by the irreversible conversion of usable forms of energy (such as mechanical, electrical, or chemical) into unusable thermal energy~\citep{Borgnakke2024Fundamentals11,schroeder2020introduction}. Landauer~\citep{landauer1961irreversibility} extended this concept into the Physics of Information. The Landauer's Principle dictates that any irreversible information operation must be accompanied by a minimum energy dissipation. The dissipation defined here is an inherent and unavoidable cost in scientific discovery. The process of navigating a high-entropy space where the agent moves from many uncertain potential solutions to a single determined solution is necessarily dissipative, because this transition constitutes an irreversible information operation. According to Landauer's Principle, the degree of dissipation is directly proportional to the number of non-solution paths that must be explored and subsequently discarded to find a valid one. This trial-and-error cost is a fundamental challenge in scientific discovery.

Using these properties as an analytical lens, we can systematically analyze the transformation of different types of information as they flow through the autonomous discovery process.

\begin{figure*}[t] 
    \centering 
    \includegraphics[width=0.8\textwidth]{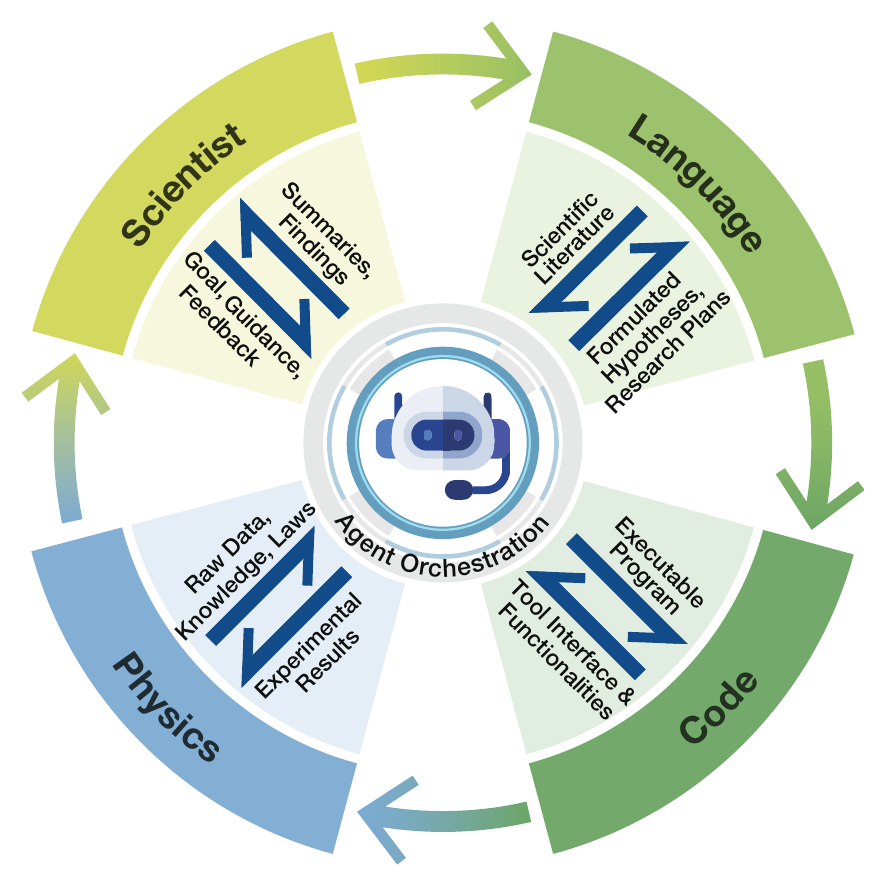} 
    \caption{An overview of autonomous agents for scientific discovery in which an agent orchestrates scientists, language, code, and physics. This figure illustrates the dynamic, closed-loop workflow of an LLM-based scientific agent as the coordinator orchestrating four key components, including scientists, language, code, and physics. The agent continuously interacts with human scientists, receiving goals, guidance, and feedback to direct research, and providing summaries and findings. Its interaction with language involves extracting knowledge from literature to formulate verifiable hypotheses and detailed research plans. The agent's interface with code translates high-level plans into executable programs for simulations or instrument control by integrating tool functionalities. Finally, it interacts with physics by using raw data and laws to direct physical or simulated instruments, yielding experimental results. This cycle represents an iterative and autonomous discovery cycle, bridging human intent to empirical evidence.}
    \label{fig:AgentOrchestration} 
\end{figure*}

\subsubsection{A Taxonomy of Information in Autonomous Scientific Discovery}

In autonomous scientific discovery, information transforms through a hierarchical process, moving from high-entropy human intent to low-entropy, verifiable physical information. Figure \ref{fig:InformationLevel} illustrates a series of transitions: from abstract Human Intent to structured Natural Language, then to formal Computer Language, and finally to verifiable Physical Information. Each step reduces entropy by resolving ambiguity and increases verifiability by making the information more precise and testable against reality. \ref{fig:AgentOrchestration} illustrates the dynamic, closed-loop workflow of an LLM-based scientific agent as the coordinator orchestrating four other key components, including Scientists, Language, Code, and Physics. These four components serve as concrete instantiations of the four corresponding information levels: Human Intent, Natural Language, Computer Language, and Physical Information.

\textbf{Human Intent} serves as the origin point of the scientific process. It exists as an abstract hypothesis or goal within a vast, unstructured conceptual space. This state possesses the highest possible Information Entropy, representing the profound uncertainty at the outset of any novel inquiry. As a pure, unarticulated concept, it is not a falsifiable proposition about the world; therefore, its Verifiability is almost zero.

\textbf{Natural Language} is the primary medium for structuring this high-entropy intent into a more tractable form. The properties of a natural language object are highly context-dependent. During creative phases like hypothesis generation, its Entropy is extremely high to allow for exploration. During planning or command phases, an agent's task is to generate a low-entropy, unambiguous statement. The Verifiability of a natural language artifact, such as a research plan, is based on its logical consistency and can be assessed against an existing knowledge base.

\textbf{Computer Language} is the medium for formal, unambiguous representation and execution. The computer language discussed in this context is primarily presented as code. Generating a computer language object, such as a query, a simulation script, or a new tool, is a process of significant entropy reduction because it transforms the ambiguity of natural language into a deterministic form. A computer language artifact possesses a high degree of Verifiability; it can be formally tested against a specification to determine if its behavior is correct.

\textbf{Physical Information} is the raw, empirical data gathered directly from the physical world. It may include measurements, signals, and experimental observations, etc. This is different from physics information, which refers to theoretical knowledge and concepts within the discipline of physics itself. Such data plays a dual role in scientific discoveries in the sense that existing datasets serve as an input for generating hypotheses, while new experimental data provides the output for testing them. Although the physical process generating the data may have high entropy due to factors like random noise or complexity, the recorded data itself is a factual observation. Consequently, physical information possesses the highest degree of Verifiability, which is formally determined by statistical significance and reproducibility. This verifiable data constitutes the evidence required to evaluate the original hypothesis.

\begin{table}[t!]
\centering
\caption{Information Analysis Across Autonomous Scientific Discovery Phases}
\label{tab:entropy_difficulty_analysis}
\resizebox{\textwidth}{!}{%
\begin{tabular}{|p{2.5cm}|p{3.2cm}|p{3.2cm}|p{3.2cm}|p{3.2cm}|p{3.2cm}|}
\hline
\rowcolor{headergray}
\textbf{Scientific Phase} & \textbf{Human Intent} & \textbf{Natural Language} & \textbf{Computer Language} & \textbf{Physical Information} & \textbf{Overall Dissipation} \\
\hline
\rowcolor{rowblue}
\textbf{Hypothesis Discovery} & \textbf{Very High Entropy} \newline The intent is abstract and completely open-ended (e.g., "discover a novel research gap"). & \textbf{Very High Entropy} \newline Requires the highest level of linguistic creativity to synthesize disparate concepts and articulate new ideas. & \textbf{Medium Entropy} \newline Primarily used for auxiliary tasks like knowledge graph construction or data mining, not the core creative act. & \textbf{High Entropy} \newline Historical data serves as inspiration; the volume is vast, unstructured, and the valuable signals are non-obvious. & \textbf{High} \newline Requires overcoming high entropy in creative thought, making it difficult to automate. \\
\hline
\rowcolor{rowgreen}
\textbf{Experimental Design} & \textbf{High Entropy} \newline The intent is creative and requires navigating a vast space of possibilities (e.g., "design a new experiment"). & \textbf{Medium Entropy} \newline Involves creating a novel, precise, and unambiguous protocol from an abstract idea. & \textbf{Medium Entropy} \newline Requires selecting, configuring, and integrating from a vast library of existing tools. & \textbf{Medium Entropy} \newline Must predict and model real-world physical laws and constraints, which are inherently complex and uncertain. & \textbf{Medium} \newline Connects abstract hypotheses to the physical world, navigating a vast and constrained search space. \\
\hline
\rowcolor{rowyellow}
\textbf{Tool Use} & \textbf{Low Entropy} \newline The intent is highly specific and direct (e.g., "run this specific analysis tool"). & \textbf{Low Entropy} \newline Typically consists of structured commands or simple instructions with low ambiguity. & \textbf{Very Low Entropy} \newline Involves calling predefined functions or APIs; the syntax and logic are deterministic. & \textbf{Low Entropy} \newline Input and output data formats are generally well-defined and structured, reducing uncertainty. & \textbf{Very Low} \newline The process is deterministic with low entropy, making it the easiest stage for AI to automate reliably. \\
\hline
\rowcolor{roworange}
\textbf{Tool Creation} & \textbf{Very High Entropy} \newline Requires defining and solving a problem for which no solution currently exists. & \textbf{Very High Entropy} \newline Needs to describe the principles, architecture, and logic of a novel, non-existent entity. & \textbf{Very High Entropy} \newline Involves designing and implementing new, complex algorithms and systems from scratch. & \textbf{Very High Entropy} \newline The new tool must reliably and robustly interact with the complex physical or computational world. & \textbf{Very High} \newline Combines the creative difficulty of hypothesis discovery with the engineering challenge of building a robust, functional system. \\
\hline
\rowcolor{rowpurple}
\textbf{Results Analysis \& Refinement} & \textbf{Low Entropy} \newline The intent is more open than tool use (e.g., "find meaningful patterns in the data"). & \textbf{Low Entropy} \newline Requires generating interpretations, summaries, and insights, allowing for more linguistic flexibility. & \textbf{Low Entropy} \newline Involves writing analysis scripts that are more complex than simple API calls but often rely on standard libraries. & \textbf{Low Entropy} \newline Raw experimental data often contains noise and artifacts, introducing uncertainty into the interpretation process. & \textbf{Low} \newline The task is constrained by the provided data, but requires sophisticated reasoning to interpret results and suggest refinements. \\
\hline
\end{tabular}%
}
\label{tab:Information Analysis}
\vspace{-1em}
\end{table}

\subsubsection{An Analysis of Information Across Autonomous Scientific Discovery Phases}

\begin{figure}[t!]
    \centering
    \includegraphics[width=0.9\textwidth]{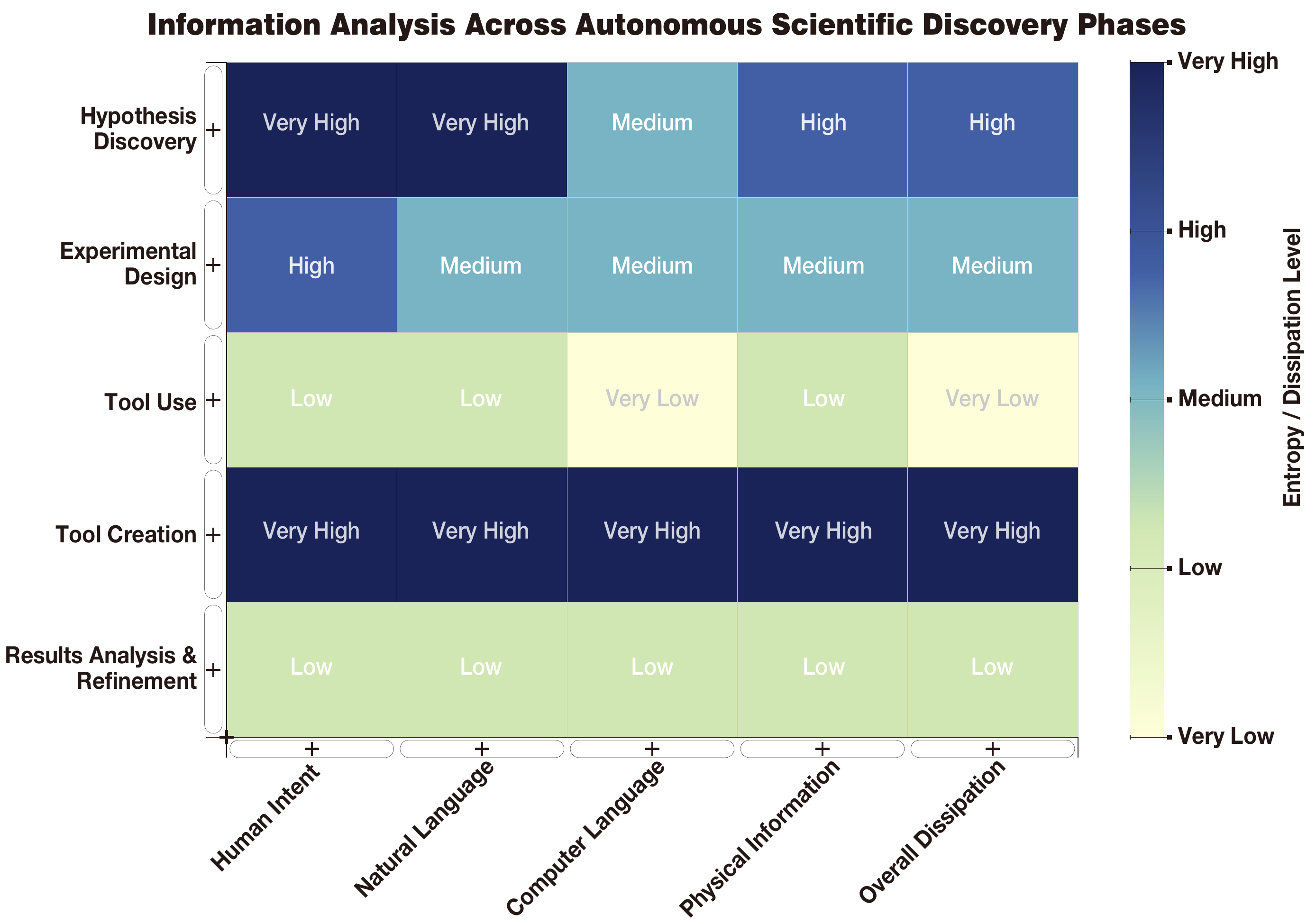}
    \caption{A heatmap representation of information analysis across autonomous scientific discovery phases.}
    \label{fig:heatmap}
\end{figure}

\begin{figure}[t!]
    \centering
    \includegraphics[width=\textwidth]{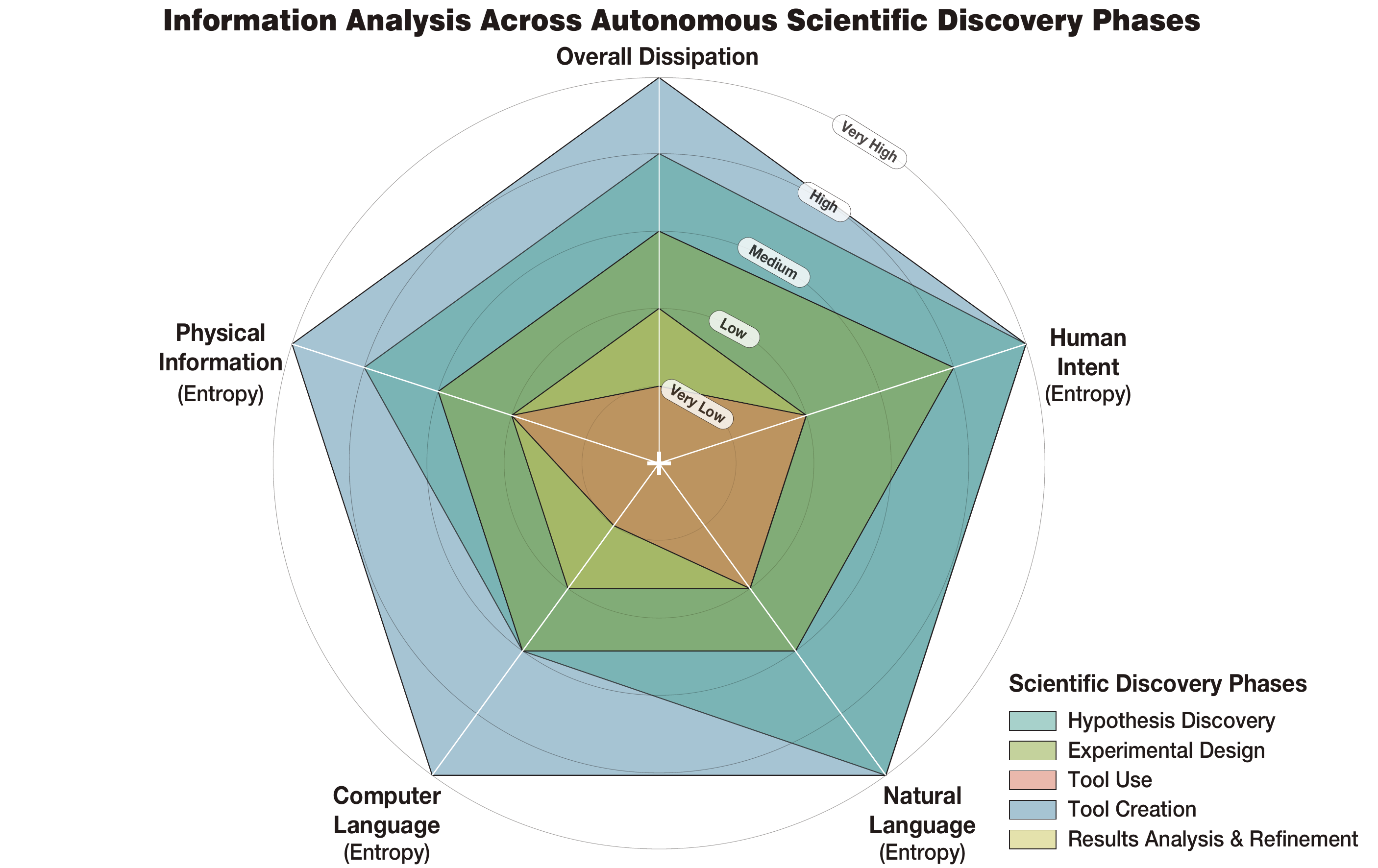}
    \caption{A radar chart representation of information analysis across autonomous scientific discovery phases.}
    \label{fig:radar}
\end{figure}

Using an information theory framework, we evaluate the entropy of different types of information, including human intent, natural language, computer language, and physical information within each phase, linking this to the difficulty of automation (dissipation). As shown in Table \ref{tab:Information Analysis} and Figure \ref{fig:heatmap} and \ref{fig:radar}, our findings reveal which phases are the most challenging to automate and which are the easiest. We will delve into the unique challenges and information characteristics of each phase, providing a blueprint for building a fully autonomous scientific discovery system.

\textbf{Hypothesis Discovery} is characterized by high levels of uncertainty, exhibiting very high entropy in Human Intent and Natural Language. The initial intent is abstract and open-ended, requiring significant linguistic creativity to articulate novel concepts by synthesizing disparate ideas. Although the computer language used for auxiliary data mining is less complex, the reliance on vast, unstructured historical data also contributes to high entropy. The immense initial uncertainty and the need to sift through a vast space of possibilities to find a single valuable insight lead to high overall dissipation, making this a fundamentally creative and challenging phase to automate.

\textbf{Experimental Design} serves as a crucial bridge from abstract concepts to concrete actions. The entropy across all types of information is medium to high. The creative intent to design a novel experiment from numerous possibilities and the need to translate that into a precise natural language protocol contribute to this uncertainty. Furthermore, orchestrating existing computational tools for simulation and modeling the inherent uncertainties of real-world physical laws adds to the complexity. This effort to structure an abstract idea into a concrete plan involves navigating a large but constrained search space, leading to medium overall dissipation.

\textbf{Tool Use} represents the most structured and deterministic phase of the scientific workflow. It is defined by low or very low entropy across all information categories. The intent is highly specific, the natural language used is typically a simple command, and the computer language required is often a direct, deterministic API call or function. Because the process is predictable and the information involved is unambiguous, the overall dissipation is very low, making this the most straightforward phase to automate reliably.

\textbf{Tool Creation} stands as the pinnacle of complexity in this framework, exhibiting very high entropy across every information category. The process begins with the highly abstract intent to solve a problem for which no tool currently exists. It requires immense creativity to describe the principles of a novel system in natural language and to design and implement new, complex algorithms in computer language. The resulting tool must also reliably interact with the complex physical or computational world. This combination of high-entropy creative demands and rigorous engineering challenges results in very high overall dissipation, marking it as the most difficult phase to fully automate.

\textbf{Results Analysis and Refinement} is a phase that operates under the strong constraints of the data generated from experiments. This grounding in data leads to low entropy across the information types involved. The intent is to find patterns within a defined dataset, and the natural language and computer language required are for interpreting and processing this data, not for open-ended creation. While this phase requires sophisticated reasoning to generate insights and suggest refinements, the task is one of inference within a bounded context rather than unconstrained exploration. This grounding in specific data results in low overall dissipation.

\subsection{Different Levels of Autonomous Agents for Scientific Discovery}

\begin{figure*}[t] 
    \centering 
    \includegraphics[width=\textwidth]{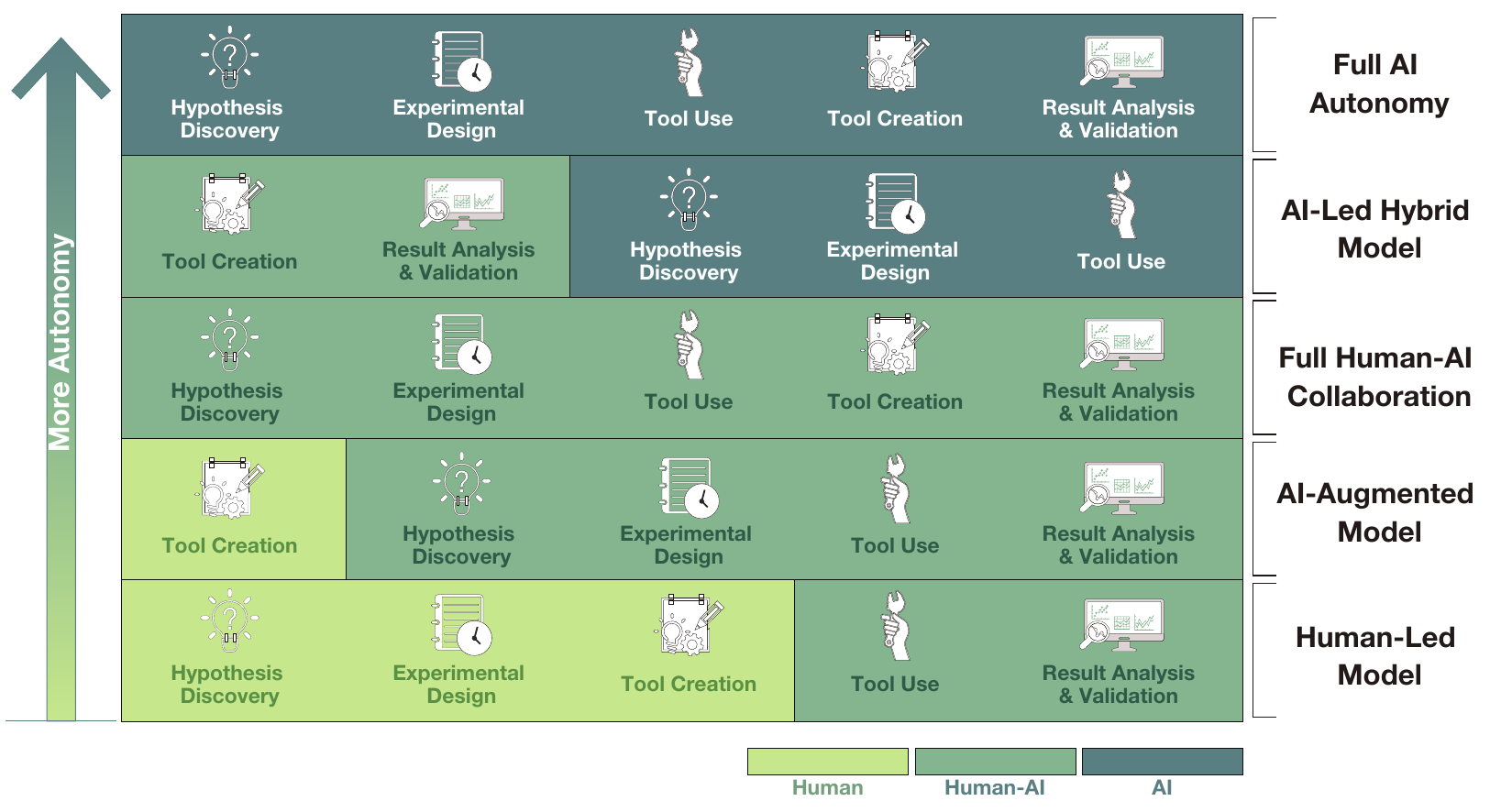} 
    \caption{A five-level framework for classifying the autonomy of scientific agents. The framework progresses from a Human-Led Model (Level 1), where the agent is confined to simple, low-entropy tasks, to an AI-Augmented Model (Level 2) and a Full Human-AI Collaboration Model (Level 3), where the agent's role expands to that of a supportive partner. A significant shift occurs in the AI-Led Hybrid Model (Level 4), where the agent takes the lead on most core scientific tasks, with humans collaborating on the most complex challenges. The final stage is Full AI Autonomy (Level 5), where the agent can manage the entire scientific process from abstract ideation to validated discovery, functioning as a truly autonomous researcher.} 
    \label{fig:AgentLevel} 
\vspace{-1em}
\end{figure*}

Figure~\ref{fig:AgentLevel} illustrates a five-level framework for scientific agents, detailing a clear progression from human-centric research to full AI autonomy across five key dimensions: 
Hypothesis Discovery, Experimental Design, Tool Use, Tool Creation, and Analysis and Refinement.

The proposed five-level framework provides a more granular and clearly delineated model of AI autonomy in science compared to existing role-based classifications. Frameworks such as those by \citet{zheng2025automation,gao2024empowering,wei2025ai} define levels based on the holistic role or overall capability of the AI system, making the boundaries between stages descriptive rather than precise.

In contrast, this new framework's strength lies in grounding the progression of autonomy in the core information-processing challenges of scientific discovery. The transition through the levels can be understood as an agent's increasing capacity to independently reduce information entropy, minimize dissipation, and systematically generate artifacts with high verifiability. This provides a more structured and measurable methodology for assessing an agent's true capabilities.

\textbf{Level 1: Human-Led Model.}
At this foundational level, the agent's role is purely instrumental, confined to the lowest-entropy and lowest-dissipation tasks where the workflow is predictable (e.g., Tool Use). Humans bear the full cognitive load of managing the immense information entropy inherent in creative processes like Hypothesis Discovery and Tool Creation. The human is the sole engine for reducing this initial uncertainty and driving the process toward a verifiable outcome.

\textbf{Level 2: AI-Augmented Model.}
The agent evolves into a supportive partner, beginning to assist in higher-entropy domains. While humans still lead, the AI collaborates in Hypothesis Discovery and Experimental Design, helping to explore the large possibility spaces. However, it relies on human guidance to navigate the high dissipation (the cost of exploring unproductive paths) and converge on a viable research direction.

\textbf{Level 3: Full Human-AI Collaboration Model.}
This level is characterized by a comprehensive partnership where every task is performed collaboratively. The agent is now capable of functioning effectively across the entire entropy spectrum, from creative ideation to data analysis. It acts as a force multiplier, jointly tackling high-entropy challenges with the human. However, it does not yet possess the autonomy to lead the process or manage the most uncertain, dissipative phases on its own.

\textbf{Level 4: AI-Led Hybrid Model.}
A critical threshold is crossed as agents take lead in autonomously reducing entropy for core scientific tasks. It independently drives Hypothesis Discovery, Experimental Design, and Tool Use, demonstrating a sophisticated ability to navigate vast and unstructured information spaces. Human collaboration remains essential for the task with the absolute highest entropy and dissipation (Tool Creation) and for the final step of analysis and refinement, where human judgment ensures the ultimate verifiability of findings.

\textbf{Level 5: Full AI Autonomy Model.}
This represents the pinnacle of autonomous science. The agent possesses complete mastery over the entire information lifecycle. A Level 5 agent can independently manage the workflow from start to finish: confronting the maximum information entropy of an abstract scientific goal, navigating the extreme dissipation inherent in creating novel tools from scratch, and producing validated, high verifiability scientific facts. At this stage, the agent operates as a truly autonomous scientific researcher.

By defining the levels based on how an agent handles these fundamental properties of information, this framework moves beyond coarse role descriptions. The boundary between levels is marked by a concrete, observable shift in the agent's ability to process uncertainty and create knowledge, offering a more robust and detailed methodology for measuring progress in the field of autonomous science.

\section{Hypothesis Discovery}

\begin{figure*}[t] 
    \centering 
    \includegraphics[width=\textwidth]{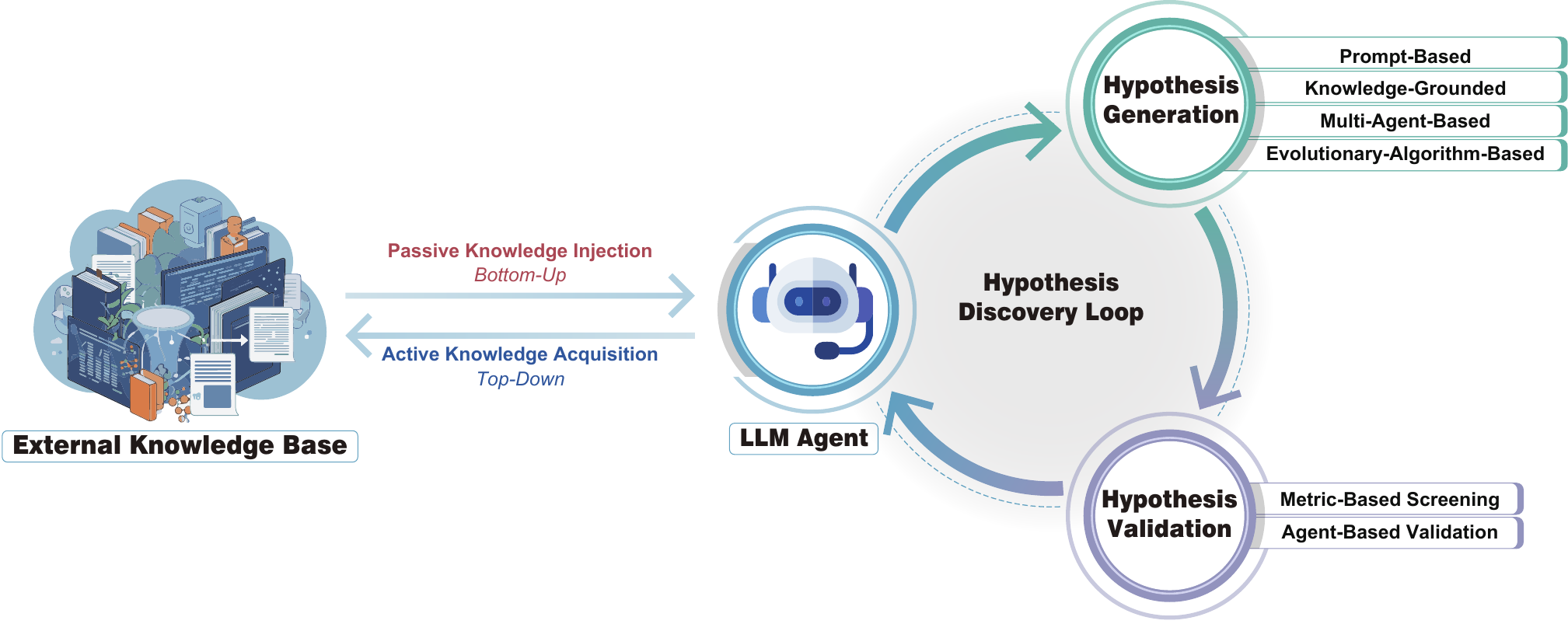} 
    \caption{The LLM-agent-driven workflow for Automated Hypothesis Discovery, centered around an iterative loop of generation and validation. The agent's reasoning is informed by an external knowledge base through two distinct paradigms: a passive, bottom-up knowledge injection and an active, top-down knowledge acquisition. During the Hypothesis Generation phase, the agent employs one of four primary strategies: Prompt-Based, Knowledge-Grounded, Multi-Agent-Based, or Evolutionary-Algorithm-Based. The resulting ideas are then filtered and refined in the Hypothesis Validation phase, which utilizes either automated Metric-Based Screening or interactive Agent-Based Validation to produce a final set of verified hypotheses.}
    \label{fig:hypothesis} 
\end{figure*}

Hypothesis Discovery is a crucial creative process in scientific discovery, with the core objective of identifying and forming novel, verifiable scientific hypotheses from vast amounts of data and existing knowledge. This process aims to accelerate the pace of knowledge discovery by revealing hidden connections that are difficult for human researchers to find alone \citep{swanson1986undiscovered}. With the rise of Large Language Models (LLMs), the level of automation and efficiency in hypothesis discovery has been enhanced to an unprecedented degree\citep{he2025reasoning,luo2025llm4sr}. A complete hypothesis discovery workflow typically includes the following three key stages: Knowledge Extraction, Hypothesis Generation, and Hypothesis Screening and Validation (Figure \ref{fig:hypothesis}).

\subsection{Knowledge Extraction}
Knowledge Extraction is the cornerstone of the hypothesis discovery pipeline. Its core task is to automatically identify, extract, and structure key information from large volumes of unstructured data, especially scientific literature, to lay the groundwork for subsequent hypothesis generation and validation. 

Traditional knowledge extraction methods involve a labor-intensive, bottom-up pipeline using techniques like Named Entity Recognition (NER) and Relation Extraction (RE) to build static, structured Knowledge Graphs (KGs) from unstructured text \citep{nickel2015review, ji2021survey}. In contrast, modern paradigms leverage the vast implicit knowledge already embedded within Large Language Models (LLMs). This shift moves away from building fixed knowledge bases and instead focuses on dynamically activating and applying the LLM's built-in reasoning capabilities. Crucially, this transition reframes the entire process: multi-step extraction tasks are now often reshaped into a single, end-to-end generative task, where the LLM directly generates the desired structured knowledge.

\textbf{Scientific Foundation Models.} To overcome the limitations of general models in specialized fields, expert models are created by training or fine-tuning them on vast, domain-specific scientific data~\citep{gururangan2020dont,pyzer2025foundation,zhao2025developing,subramanian2023towards,barman2025large,narayanan2025training}. For instance, models like BioBERT are pre-trained on biomedical texts, enabling them to more accurately extract specific entities like genes, diseases, and chemicals, as well as the relationships between them from research articles~\citep{lee2020biobert}. Similarly, generative models like BioGPT can not only extract but also summarize complex biological information coherently~\citep{luo2022biogpt}. On a broader scale, models such as Galactica were designed with the explicit goal of structuring and organizing scientific knowledge from papers, lecture notes, and textbooks, essentially performing knowledge extraction on a massive scale~\citep{taylor2022galactica}. By internalizing the nuances of a specific scientific language, these models serve as far more effective tools for building accurate and comprehensive knowledge bases from technical documents.

\textbf{Retrieval-Augmented Generation (RAG).} Retrieval-Augmented Generation (RAG) is a technique designed to address the critical issues of outdated knowledge and hallucination in LLMs by connecting the model to external knowledge sources~\citep{lewis2020retrieval, guu2020retrieval, gao2023retrieval, izacard2023atlas, fan2024survey}. In the context of knowledge extraction, its primary function is to transform vast, unstructured scientific literature into a verified and structured knowledge base~\citep{garcia2024review, lopez2025clinical, feng2025retrieval, krotkov2025nanostructured}. By retrieving relevant documents and using them as a direct source, RAG generates factually-grounded and traceable outputs, such as precise summaries or extracted entity-relationship pairs, which serve as the reliable foundation for subsequent analysis and discovery~\citep{lee2025rag,xie2024cassia,gao2025frag}.

\textbf{Multimodal Knowledge Extraction.} With the advent of multimodal models and agents that can invoke analysis tools, knowledge extraction from scientific literature is no longer confined to text~\citep{khalighinejad2024matvix, odobesku2025agent}. For instance, systems like ChemMiner leverage multimodal capabilities to extract chemical information from both text and diagrams within papers~\citep{chen2024autonomous}, while ChartAssistant demonstrates the ability to reverse-engineer scientific charts by converting visual data back into structured tables~\citep{meng2024chartassisstant}. Similarly, the use of Multimodal Models, such as GPT-4o, enables the processing of complex biomedical documents by integrating information across text, tables, and diagrams to answer questions~\citep{hurst2024gpt}.

\subsection{Hypothesis Generation} 
Hypothesis generation is the core of hypothesis discovery, aiming to formulate plausible and testable statements that explain observed phenomena or predict outcomes under certain conditions. In scientific practice, it is a creative yet structured step that bridges observation and prediction. A good hypothesis should be grounded in existing knowledge and make predictions that can be empirically tested and potentially proven wrong. Traditionally, scientists generate hypotheses by identifying gaps or contradictions from previous literature, drawing analogies from other domains, and applying domain expertise to reason. Recent advancements are progressing beyond classic literature-based discovery (LBD) ~\citep{swanson1986undiscovered} to develop dynamic, agentic systems capable of independently reading, reasoning, and iteratively proposing sophisticated, testable conjectures. This evolution in methodology can be categorized into four key approaches:

\textbf{Prompt-Based Systems.} At its most fundamental level, hypothesis generation using LLMs can be achieved through prompt-based methods. This approach directly leverages the extensive knowledge an LLM acquires during its pre-training phase. By formulating a carefully constructed prompt, a user can query the model to propose hypotheses in a zero-shot fashion~\citep{qi2023large, zhou2024hypothesis}. Early evidence shows LLMs can act as zero-shot hypothesis proposers in biomedicine~\citep{qi2023large}, relying on their pre-trained knowledge alone to formulate ideas. The efficacy of this method is contingent upon the LLM's internal representation of scientific concepts and relationships. While this allows for the exploration of a broad range of ideas, it also introduces a susceptibility to generating plausible but factually incorrect statements or restating known facts~\citep{xiong2025toward}.

\textbf{Knowledge-Grounded Systems.} To enhance the factual accuracy and novelty of generated hypotheses, knowledge-grounded methods have been developed. These techniques augment the LLM's internal knowledge with external, authoritative information. One prominent technique is Retrieval-Augmented Generation (RAG), which dynamically retrieves unstructured textual evidence from literature. By synthesizing information from multiple, potentially disparate sources, it enables the model to infer novel connections and propose innovative, evidence-backed hypotheses~\citep{bazgir2025proteinhypothesis}. Another powerful technique involves the injection of structured knowledge, often from curated knowledge graphs (KGs). KG-CoI~\citep{xiong2024improving}, for example, utilizes structured knowledge sources such as knowledge graphs (KGs) to ground its hypotheses, improving the accuracy of its reasoning chains and reducing factual errors. ResearchLink \citep{borrego2025research}, for example, integrates graph embeddings, path-based features, and bibliometric data to generate cross-domain hypotheses with high precision. Moreover, a recent study \citep{tong2024automating} presents a novel framework that combines large language models (LLMs) with causal knowledge graphs to generate psychological hypotheses. By analyzing over 43,000 psychology articles, the approach produced hypotheses on well-being that matched expert-level novelty and significantly outperformed LLM-only methods. 

\textbf{Multi-Agent Systems.} To broaden the scope of hypothesis generation, multi-agent frameworks emulate collective brainstorming by leveraging diversity from multiple sources~\citep{wu2023autogen,chase_2022_langchain}. This approach generates a richer pool of initial ideas by either utilizing different foundation models as distinct agents or by assigning different roles and prompts to instances of the same model. These varied configurations steer the models to generate outputs from different conceptual angles by activating distinct generative distributions. By synthesizing ideas from these diverse agents, the framework effectively uncovers less obvious connections and produces a more comprehensive set of potential hypotheses than a monolithic approach could achieve alone~\citep{ chen2025beyond,bazgir2025proteinhypothesis}. For example, systems such as AI co-scientist \citep{gottweis2025aicoscientist} employ a proposer-critic dynamic, where some agents generate initial ideas while others rigorously challenge their assumptions to refine them through interaction. This paradigm is implemented in specialized frameworks like ACCELMAT for materials science \citep{kumbhar2025hypothesisgenerationmaterialsdiscovery}, which uses a structured, iterative loop of proposal and critique among multiple agents to progressively enhance the quality of novel material hypotheses. VIRSCI \citep{su2024many} extends this by simulating scientific teams using real-world academic data, enabling agents to form collaborative research teams and generate novel ideas through inter- and intra-team discussion mechanisms. Similarly, AstroAgents \citep{saeedi2025astroagents} deploys domain-specific agents to interpret mass spectrometry data and hypothesize about prebiotic chemical pathways related to the origins of life, with more than 30\% of hypotheses validated as scientifically plausible by expert reviewers.

\textbf{Evolutionary Algorithm-Based Systems.} For a dynamic and automated LLM-based approach to discovery, closed-loop systems apply principles from evolutionary algorithms to treat hypothesis generation as an optimization problem. An initial population of hypotheses undergoes iterative refinement using operators such as mutation and recombination, while a fitness function evaluates the quality of the hypothesis to guide the selection. This paradigm is exemplified by systems like MOOSE-Chem~\citep{yang2025moosechemlargelanguagemodels}, which uses evolutionary search to guide different ways of associating inspirations and background to generate hypotheses. This combination ensures that generated hypotheses are not only optimized but also validated for novelty against the knowledge baseline established during the LLM's pretraining. Similarly, HypoAgents \citep{duan2025bayes} integrate Bayesian updating within an evolutionary loop: agents propose hypotheses, test them via RAG-informed evidence, update probabilities, and refine uncertain hypotheses in a continuous scientific feedback cycle. MolLEO \citep{wang2024efficient} leverages LLMs directly as mutation and recombination operators, prompted with text-based instructions and desirable target properties. LLM-based evolutionary systems show immense promise for navigating vast combinatorial spaces \citep{liu2025large} with textual information extraction or rich, semantic objectives, such as discovering new materials \citep{jia2024llmatdesign}, mechanical structures \citep{jadhav2024large}, molecules \citep{wang2024efficient} or macromolecules \citep{reinhart2024large}. They also show potential for multi-objective numerical optimization \citep{liu2024large}, which includes real-world problems like optimizing nozzle design, heat transfer or wind farm layout \citep{brahmachary2025large}.

\subsection{Hypothesis Screening and Validation}
Rigorous screening and validation of automatically generated hypotheses rely on a clear set of evaluation axes, such as plausibility, novelty, feasibility, and cost~\citep{alkan2025survey, bazgir2025agentichypothesis}. This step is necessary to manage the large volume of generated ideas and requires transparent assessment methods to avoid over-claiming~\citep{beel2025evaluating}. Methodologies for this stage can be broadly classified into two approaches: automated evaluation based on defined metrics and dynamic refinement through agent-based critique. 

\textbf{Metric-Based Screening.} This approach focuses on developing automated, scalable, and objective criteria to evaluate hypotheses. Some frameworks implement formal statistical controls; for example, POPPER introduces an automated validation framework that achieves human-level accuracy at significantly greater speed by applying rigorous statistical checks~\citep{huang2025automatedhypothesisvalidationagentic}. Other methods focus on specific metrics like novelty, where systems such as SCIMON explicitly compare new ideas against existing literature and revise them until they no longer resemble prior work~\citep{wang2024scimon}. To enhance external validity and enable consistent comparisons, standardized benchmarks are emerging. These include task-specific suites like MATDESIGN for materials science, which introduces scalable quality metrics~\citep{kumbhar2025hypothesis}, and general platforms like HypoBench, which provides a main benchmark for hypothesis discovery across diverse tasks~\citep{liu2025hypobench}. Beyond these, recent approaches employ LLM-derived priors for automatic hypothesis assessment. In particular, the Logit-based Calibrated Prior technique \citep{gong2025exploiting} uses LLM-generated expectations to quantify how surprising a correlation is, enabling ranking of hypotheses by novelty and relevance with high accuracy in real-world data. Meanwhile, Bayesian frameworks have been deployed for evaluating model capabilities \citep{xiao2025confidence}, treating performance assessment as a hypothesis testing task under uncertainty—an approach that introduces probabilistic robustness when sample sizes are limited.

\textbf{Agent-Based Validation.} This approach simulates the collaborative and critical processes of scientific inquiry, where agents interact to challenge and improve hypotheses. Frameworks like ResearchAgent facilitate an iterative dialogue between a proposal-generating agent and a panel of reviewer agents. This process mimics peer review, refining the hypothesis until it meets rigorous criteria~\citep{baek2024researchagent}. Similarly, to ensure empirical testability, other systems close the loop between theory and evidence. The Scientific Generative Agent (SGA), for instance, combines an LLM acting as a theorist with scientific simulators that function as experimental systems. This allows the agent to immediately test and revise its hypotheses based on simulated experimental outcomes, ensuring ongoing refinement based on feedback~\citep{ma2024llm}.

\begin{figure*}[t] 
    \centering 
    \includegraphics[width=\textwidth]{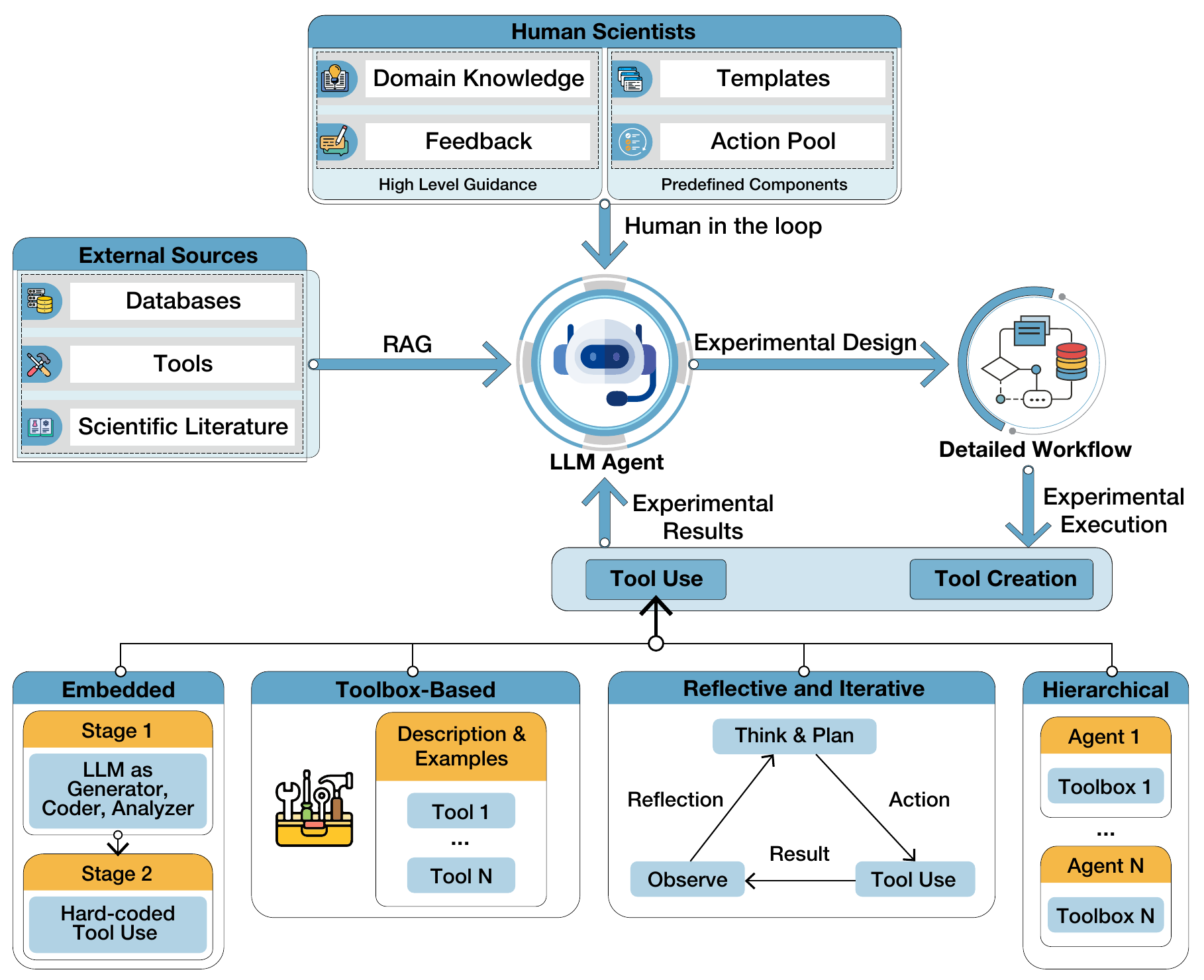} 
    \caption{An LLM-agent-driven workflow of automated Experimental Design and Execution. This workflow illustrates how an LLM agent orchestrates experimental design and execution. In the Experimental Design phase, the central agent generates a detailed workflow by integrating information from multiple inputs. It receives high-level guidance (domain knowledge, feedback) and predefined components (templates, action pools) from Human Scientists, leverages Retrieval-Augmented Generation (RAG) to pull in knowledge from External Sources, and incorporates Experimental Results from previous execution cycles as a critical feedback mechanism for refinement. For the Experimental Execution phase, the agent acts upon the detailed workflow, deciding between using existing instruments (Tool Use) or developing new ones (Tool Creation). Tool use can be implemented through four primary strategies for tool interaction: Embedded, Toolbox-Based, Reflective and Iterative, and Hierarchical.} 
    \label{fig:Experiment} 
\end{figure*}

\section{Experimental Design and Execution}

In this section, we analyze the critical phases of Experimental Design and Execution within the autonomous scientific discovery process. As depicted in Figure \ref{fig:Experiment}, we first examine how LLM agents generate robust experimental plans by integrating high-level human guidance, leveraging external knowledge sources via Retrieval-Augmented Generation (RAG), and incorporating feedback from prior experiments. We then detail the Experimental Execution phase, focusing on the various strategies, from highly structured to dynamically collaborative, that agents use to interact with and manage external tools, bridging the gap between abstract plans and physical reality.

\subsection{Experimental Design}
A critical step in scientific discovery, following hypothesis generation, is experimental design. This involves creating a structured plan to systematically test ideas and achieve research objectives. The process is fundamentally a form of workflow generation, where high-level scientific goals are translated into concrete, executable protocols. LLM-based scientific agents are increasingly used for this task, combining natural language reasoning with planning capabilities, often enhanced by retrieval tools and physical knowledge, to generate coherent and detailed experimental plans.

The core challenge for LLM agents in experimental design is navigating a nearly infinite space of possible action sequences to find a solution that is both scientifically sound and practically feasible. To address this, agents employ various strategies to effectively constrain and navigate this complex space.\citep{huang2025biomni, zhou2025greaterautonomymaterialsdiscovery, cao2024agents, yu2024chemtoolagent} These strategies can be grouped into four main categories:  RAG for Grounded Planning, Human High-Level Guidance, Templates and Predefined Actions, Post-Execution Feedback.

\textbf{Retrieval-Augmented Generation for Grounded Design.}
This strategy ensures that an agent's plan is scientifically valid from the outset by grounding its reasoning in external, reliable knowledge sources. Instead of relying solely on its internal training, the agent uses a Retrieval-Augmented Generation (RAG) approach to query specialized tools, databases, and scientific literature. This allows the agent to build its experimental plan on a foundation of established facts and methods. 

Biomni~\citep{huang2025biomni} first creates a comprehensive action space by automatically mining tens of thousands of scientific papers for relevant tools and databases. When given a task, it retrieves the most pertinent tools from this space to construct its plan. TxAgent~\citep{gao2025txagent} uses a specialized retrieval model called ToolRAG to select the most appropriate tools from its ToolUniverse of 211 clinical instruments, ensuring its therapeutic reasoning is based on evidence. Chemist-X~\citep{chen2023chemist} leverages RAG to search literature and molecular databases to define a promising and constrained search space for reaction condition optimization before any physical experiments are run. CLADD \citep{lee2025rag} has a Planning Team which retrieves the most relevant resources from databases and tools. The Molecule Understanding Team then generate structured reports based on these external sources which are synthesized by a Prediction Agent.

\textbf{Human High-Level Guidance.}
For complex or novel scientific challenges, LLMs may struggle to generate a high-quality experimental design on its own. As noted in some studies \citep{zhou2025greaterautonomymaterialsdiscovery,huang2025crispr}, human expertise is often crucial to refining and validating the agent plan. In this collaborative model, the human scientist acts as a supervisor, providing high-level guidance and feedback to guide the agent's decision-making process.

MAPPS~\citep{zhou2025greaterautonomymaterialsdiscovery} framework includes a Scientific Mediator module. This component allows human scientists to interact with the agent, providing feedback, offering guidance, and making corrections to the experimental plan, ensuring that the agent's autonomous discovery process remains aligned with scientific principles and the researcher's goals. AI co-scientist \citep{gottweis2025aicoscientist} introduces a scientist-in-the-loop paradigm, where human scientists actively collaborate with agents throughout the research cycle. Human scientists interact with the system by specifying a research goal, suggesting their own ideas, providing feedback and reviews. In Virtual Lab \citep{swanson2024virtual}, The human researcher defines the Principal Investigator (PI) and Scientific Critic agents by specifying their title, expertise, goal, and role, and sets the agenda for each meeting, which guides the agents to further design experiments and solutions.

\textbf{Human-Provided Templates and Predefined Actions}
This is a more structured form of human-AI collaboration where humans define the fundamental building blocks of an experiment, and the LLM's role is to assemble them into a coherent workflow. Humans may provide a set of allowed "meta-actions" in an action pool or create high-level workflow templates. The agent then fills in the specific parameters and details. This approach often involves modeling the experiment as a state machine, where the agent's primary task is to establish the rules for transitioning between predefined experimental states.

The k-agents \citep{cao2024agents} framework exemplifies this by having an execution agent that decomposes a human-provided experimental procedure into a state machine. The agent then interacts with other knowledge agents to execute each state and uses the results to determine the next transition, enabling closed-loop control.  CRISPR-GPT~\citep{huang2025crispr} utilizes expert-predefined meta-tasks as high-level templates, while its LLM planner decomposes the user's high-level request and links it to corresponding state machines to assemble a complete, executable workflow. PerTurboAgent~\citep{hao2025perturboagent} operates using a predefined Action Pool containing high-level actions like predict, reflect, and refine. The agent designs its experiment by selecting and sequencing these actions to decide which genes to perturb in each round. 

\textbf{Post-Execution Feedback.}
This strategy does not require a perfect plan from the start. Instead, the agent's design process is an exploratory one, where it learns and adapts by observing the results of its actions. This feedback can be used to make immediate, step-by-step corrections or to inform the design of the next major experimental cycle.

Agents like ChemToolAgent \citep{yu2024chemtoolagent} use the ReAct loop to solve chemistry problems by calling a tool, observing the output, and planning the subsequent action based on that result. BioInformatics Agent~\citep{xin2024bioinformatics} uses a self-correction loop; if an analysis workflow's initial output doesn't meet user expectations, it modifies the entire workflow and re-executes it. BioDiscoveryAgent~\citep{roohani2024biodiscoveryagent} operates in a closed loop, using the results from one round of genetic perturbations to reason about which genes are most promising to test in the subsequent round. OSDA Agent \citep{hu2025osda} adopts an Actor-Evaluator-Reflection loop, where an LLM-based Actor generates candidate OSDA molecules, which are subsequently assessed by an Evaluator through chemical validity checks. A Self-reflector module then integrates this feedback to guide the iterative refinement of the Actor’s outputs.

\subsection{Experimental Execution}

The experimental execution stage is a central phase in the scientific discovery process, serving as the crucial bridge that transforms abstract strategies into the concrete actions and empirical evidence needed for generating insights. It involves the systematic implementation of a designed workflow, which extends beyond simply running code to include a range of complex tasks: orchestrating computational resources, managing large datasets, interfacing with laboratory instruments or simulators, and continuously monitoring the workflow to ensure its correctness, efficiency, and reproducibility. LLM-based agents operate at the intersection of natural language instructions, computer language, and physical information, enabling a fluid transition from concept to execution. The execution of planned experiments by LLM agents is primarily achieved through two interconnected capabilities, namely tool use and tool creation.

\subsubsection{Tool Use}
For LLM-based agents to be effective in science, the ability to use tools is essential. First, general-purpose tools like search engines allow agents to survey the dynamic landscape of scientific research. This ensures their work is grounded in the most recent findings and established principles. More critically, since LLMs are fundamentally text-based models and lack the ability to directly interact with or measure the physical world, they must rely on domain-specific tools for analysis and exploration. By interfacing with chemical simulation software, data analysis pipelines, or robot lab equipment, agents can bridge the gap between text-based reasoning and empirical science. Over the past few years, several distinct strategies for integrating agents with external tools have emerged, representing a spectrum of autonomy from highly controlled pipelines to dynamic, collaborative systems.\citep{hu2025osda, swanson2024virtual, jin2024agentmd, yu2024chemtoolagent, huang2025biomni} To provide a structured view, we categorize these strategies into four representative models of tool use: Embedded Tool Use, Toolbox-based Tool Use, Reflective \& Iterative Tool Use, and Hierarchical Tool Use in Multi-Agent Systems.

\textbf{Embedded Tool Use.}
The Embedded Tool Use method represents a structured and highly controlled approach to integrating AI with scientific workflows. In this model, developers pre-define a complete research pipeline where specific tools are hard-coded to execute at particular stages. The Large Language Model (LLM) acts not as a decision-maker for tool selection, but rather as a component within the workflow. Its role is often confined to interpreting the output of a tool, generating code for a specific, predetermined step, or translating human instructions into parameters for the next tool in the sequence. This method ensures reliability and reproducibility, making it well-suited for routine, high-throughput analyses where the scientific process is already well-established. 

For example, OSDA Agent \citep{hu2025osda} embeds the LLM into a fixed "Actor-Evaluator-Reflector" loop, where the LLM acts as a molecule generator and reflector, while a set of hard-coded computational chemistry tools serves as the evaluator to validate its output. LLMatDesign \citep{jia2024llmatdesign} adopts a pre-set iterative design loop, in which the LLM is responsible for proposing material modification plans and reflecting on the results, but the specific structure relaxation and property prediction are executed by hard-coded Machine Learning Force Field (MLFF) and Machine Learning Property Predictor (MLPP) tools. MatLLMSearch \citep{gan2025large} embeds the LLM into a fixed evolutionary algorithm framework, where its role is confined to the stage to generate new crystal structure candidates, which are then passed to a preset evaluation pipeline composed of tools like MLFF and DFT. scBaseCount \citep{youngblut2025scbasecount} embeds the LLM in a fixed pipeline for large-scale single-cell data retrieval, preprocessing, normalization and formatting, to create a unified repository for single-cell RNA-seq data. Chemist-X \citep{chen2023chemist} embeds the LLM to interact with pre-packaged API functions including CAD tools and ML models.

\textbf{Toolbox-Based Tool Use.} This method empowers the AI agent with significantly more autonomy. The agent is provided with a toolbox or a library of available functions. Each tool is accompanied by a natural language description of its purpose, inputs, and outputs. The LLM leverages its reasoning capabilities to interpret a user's high-level goal, decompose it into smaller steps, and select the most appropriate tool from the toolbox to accomplish each step. This approach transforms the LLM from a simple component into a central coordinator, capable of orchestrating complex sequences of operations. 

This approach is demonstrated in works such as AgentMD~\citep{jin2024agentmd} is a clinical agent that curates and applies medical risk calculators by selecting from 2,164 mined clinical tools. ChemCrow~\citep{m2024augmenting} bridges drug discovery and materials science using 18 expert-designed tools. scAgent \citep{mao2025scagent} contains a tool hub offers over 30 plugins that centralizes diverse single-cell analysis methods into a unified interface. TxAgent \citep{gao2025txagent} leverages a broad Tool Universe, spanning drug databases, molecular analysis, and clinical resources, which allows the agent to dynamically select and integrate multiple tools for therapeutic reasoning. GIS Copilot \citep{akinboyewa2025gis} integrates Agent into QGIS to autonomously conduct spatial analysis. Equipped with comprehensive tool documentation and external libraries, it enables the agent to dynamically dynamically select, generate, and execute geospatial analysis workflows. SciToolAgent \citep{chen2025scitoolagent} builds a Scientific Tool Knowledge Graph that maps the intricate relationships, dependencies, and compatibilities among a vast library of scientific tools , which allows an LLM to create an optimal Chain-of-Tools tailored for a specific scientific query. This graph-based toolbox overcomes the limitations of previous agents that struggled with integrating a large and diverse toolset.
  
\textbf{Reflective and Iterative Tool Use.} 
 This approach moves beyond simple tool selection, enabling the agent to reflect on the results of each tool use and dynamically plan its next steps. This process mimics the human approach to problem-solving: try, observe, reflect, and then decide what to do next. The ReAct ~\citep{yao2023react} (Reasoning and Acting) framework is a prime example of this methodology. It operates in a continuous loop of Thought-Action-Observation. Similarly, the CodeAct~\citep{wang2024executable} framework is a specialized extension of ReAct. Here, the agent's actions involve generating and executing code in a sandboxed environment. After execution, the agent observes the code's output or any error messages. If the code fails, the agent reflects on the error, iteratively generates a corrected version, and tries again. 
 
 ChemToolAgent~\citep{yu2024tooling} is a prime example, using the ReAct framework to orchestrate a suite of 29 different tools, from searching the PubChem database to predicting molecular properties. It can reason about a user's chemistry question, call the right tool, and then analyze the output to decide the next logical step, like performing another calculation or synthesizing the results into a final answer. Biomni~\citep{huang2025biomni} is a general-purpose biomedical AI agent that automates complex data analysis by leveraging the CodeAct framework. This code-based iterative method allows Biomni to navigate vast, interconnected knowledge domains with the flexibility of a virtual biologist, writing, running, observing, and debugging code to accomplish its goals. PerTurboAgent \citep{hao2025perturboagent} is an agent for sequential Perturb-seq. At each round, the agent integrates prior knowledge, perturbation-effect models, and past experimental results to reflect on outcomes, update predictions, and propose the next gene panel. This closed-loop process adaptively prioritizes perturbations with the highest phenotypic impact.

\textbf{Hierarchical Tool Use in Multi-Agent Systems.}
As the complexity of scientific inquiry increases, a single agent may struggle to manage an overwhelmingly large and diverse toolbox. The hierarchical delegation method addresses this by employing a team of specialized AI agents. A high-level planner agent first deconstructs a complex problem into sub-tasks. These sub-tasks are then delegated to different expert agents, each equipped with its own smaller, specialized toolbox. This division of labor makes the tool selection process more manageable and robust, mirroring the structure of human scientific teams.  

For example, the Team of AI-made Scientists (TAIS)~\citep{liu2024toward} framework simulates a research team by assigning LLM agents specialized roles such as planner, data handler, and analyst. Similarly, MedAgents~\citep{tang2023medagents} uses a collaborative approach where distinct medical expert agents, each with specialized tools, iterate to reach a consensus diagnosis. In chemistry, ChemAgents~\citep{song2025multiagent} employs a hierarchical system with five specialized agents, including a Literature Reader and Robot Operator. In materials, AtomAgent \citep{ghafarollahi2024atomagentsalloydesigndiscovery} collaborate across knowledge retrieval, multimodal data analysis, and molecular simulations, autonomously orchestrating alloy design workflows and accelerating data-driven discovery of high-performance materials. For protein design, ProtAgents~\citep{ghafarollahi2024protagents} deploys multiple agents with distinct skills that dynamically collaborate on engineering tasks. The ``ChatGPT Research Group''~\citep{zheng2023chatgpt} integrates seven specialized assistants for MOF/COF synthesis, with agents dedicated to planning, literature mining, and robot operation. In antiviral antibody design, Virtual Lab \citep{swanson2024virtual} introduces a multi-agent system where a PI agent coordinates scientist agents using the protein language model (PLM) ESM \citep{lin2023evolutionary}, the protein folding model AlphaFold-Multimer \citep{abramson2024accurate}, and the computational biology software Rosetta \citep{boorla2023novo}, generating 92 nanobody candidates and experimentally validating two with improved binding to SARS-CoV-2 variants. In physics, the k-agents framework~\citep{cao2024agents} enables an autonomous quantum laboratory where agents plan and execute experiments on quantum processors. 
    
\subsubsection{Tool Creation}

While many AI agents demonstrate a remarkable ability to use existing software libraries and digital tools, a new frontier is emerging where their primary function shifts from application to invention. The most advanced scientific agents are transcending mere tool orchestration to create entirely new scientific tools and algorithms. This paradigm shift marks the evolution of AI from a digital assistant that follows a playbook to a creative partner capable of contributing novel methods and solutions to complex scientific problems.

At the forefront of the creative paradigm are agents that generate code not just to execute a workflow, but to discover new methods. ToolUniverse~\citep{gao2025democratizing} is designed to democratize the power of AI scientists by enabling LLM agents to not only execute workflows with a comprehensive suite of existing tools, but also autonomously invent, implement, and integrate new, reusable scientific algorithms, transforming agents from mere users to creative toolmakers.
MAPPS~\citep{zhou2025greaterautonomymaterialsdiscovery} represents a powerful hybrid approach that blends the use of existing tools with the creation of new algorithms. Its code generation is twofold: it generates code to use established physics-based foundation models for analysis, but more importantly, it simultaneously invents and implements new algorithms tailored to discover novel materials more efficiently. CodePDE~\citep{li2025codepdeinferenceframeworkllmdriven} iteratively generates, tests, and refines code to produce high-performance solvers for partial differential equations (PDEs). The final output is not just the solution to a single problem, but a robust, reusable solver that constitutes a new and valuable scientific tool. AlphaEvolve~\citep{novikov2025alphaevolvecodingagentscientific} takes this concept further by employing an evolutionary framework. It treats a population of algorithms as a gene pool and uses an LLM to ``mutate'' the code, iteratively evolving better solutions. This process has led to the discovery of entirely new algorithms that are more efficient than those designed by humans, showcasing code generation as a mechanism for pure algorithmic invention~\citep{nagda2025reinforced}. ShinkaEvolve~\citep{lange2025shinkaevolve} similarly evolves programs, improving sample efficiency through exploration–exploitation-balanced parent selection, novelty-based program rejection sampling, and bandit-based, task-dependent LLM prioritization. TOOLMAKER~\citep{wolflein2025llm} introduces an autonomous framework for transforming scientific code repositories into LLM-compatible tools, enabling agents to not only employ but also systematically generate new utilities, with cross-domain evaluations demonstrating substantially higher reliability than prior software-engineering agents.

\section{Result Analysis and Refinement}

\subsection{Result Analysis}
Experimental result analysis in scientific discovery is the critical process where autonomous agents interpret raw outputs—such as numerical data, plots, and images—to derive meaningful scientific insights \citep{Gridach2025}. This stage presents significant challenges, including the need to handle complex, multimodal data types and ensure the analysis is scientifically robust \citep{Liu2025, Zhang2024}. The approaches to this task are classified into distinct paradigms based on the core mechanism the agent uses to interact with and process the data. 

The analysis of experimental results by AI agents can be categorized into three primary paradigms, distinguished by the agent's core mechanism for interacting with and interpreting scientific data. These approaches are not mutually exclusive; rather, the most advanced systems often blend them to tackle complex, multi-faceted research problems. Table \ref{tab:result_analysis} presents a concise overview of three distinct approaches to analyzing and interpreting scientific results.

\textbf{Modality-Driven Analysis.} This paradigm is defined by its reliance on the advanced perceptual and interpretive capabilities of Multimodal Large Language Models (MLLMs) \citep{han2023chartllama,meng2024chartassisstant,hurst2024gpt,li2024geometry,fu2024fragment,yan2024invariant}. The core mechanism involves the direct processing of non-textual data formats, such as images, charts, videos, and even audio, to extract scientific insights. Instead of relying on structured numerical inputs, the agent "sees" the data as a human researcher would. This approach is critical for disciplines where visual evidence is paramount. Representative systems like The AI Scientist-v2 \citep{yamada2025ai} leverage a Vision-Language Model (VLM) in a feedback loop to critique and refine generated plots, ensuring they are not only accurate but also clearly interpretable. Other specialized work, such as PlotGen \citep{luo2024plotgen}, focuses on deconstructing scientific charts into their semantic components—axes, legends, data points—to perform detailed analysis and verification. 

However, this paradigm faces significant limitations. The inherent heterogeneity of scientific data (e.g., diverse chart types, microscopy images, spectral graphs) poses a major processing challenge. Furthermore, the ambiguity of visual information can lead to misinterpretations, and the high computational cost of training and running large-scale MLLMs remains a practical barrier.

\textbf{Tool-Augmented Analysis.} As the most prevalent and mature paradigm, tool-augmented analysis centers on the agent's ability to interact with external, domain-specific resources, including software APIs, databases, and even physical hardware. This approach grounds the LLM's reasoning process in the deterministic and validated outputs of specialized tools, mitigating the risk of factual hallucination and enabling highly technical operations. For instance, ChemCrow~\citep{bran2023chemcrow} is equipped with a suite of chemistry tools for tasks like predicting reaction outcomes and searching molecular databases. Coscientist~\citep{boiko2023autonomous} has demonstrated the ability to not only use bioinformatics software but also control liquid handling robots to execute experiments. Similarly, Virtual Lab~\citep{boeck2023multiagent} follows a computational pipeline of specialized tools for nanobody design. 

The primary limitations of this paradigm include the reliability of the external tools themselves, the risk of error propagation through a chain of tool calls, and practical challenges like API costs, latency, and the lack of interface standardization across the scientific software ecosystem.

\textbf{Computation-Native Analysis.} This paradigm leverages the agent's intrinsic ability to generate and execute code and perform symbolic reasoning. It treats data analysis as a computational task to be solved from first principles, rather than relying on pre-built tools. This grants the agent maximum flexibility, allowing it to handle a wide array of structured data formats like tabular data, text, and numerical arrays to perform bespoke analyses. For example, Data Interpreter translates natural language queries into Python scripts, using libraries like Pandas and Matplotlib within a secure code interpreter to execute a full data science workflow from cleaning to visualization \citep{hong2024data}. In the realm of fundamental science, frameworks like LLM-SR and MOBLLM use the agent's code generation capabilities to perform symbolic regression—discovering the underlying mathematical equations that describe experimental data \citep{kamienny2024llmsr, binbas2024autonomous}. 

The key challenges here are technical and philosophical: the risk of "code hallucination" (generating incorrect or inefficient code), the difficulty of scaling logical reasoning for highly complex problems, and the fundamental question of verifying whether an agent is making a novel discovery versus merely recalling a similar solution from its vast training data.

\begin{table}[t]
\centering
\caption{Comparison of Results Analysis Paradigms}
\label{tab:paradigm_comparison_simplified}
\resizebox{\textwidth}{!}{%
\fontsize{10pt}{6pt}\selectfont
\begin{tblr}{
  colspec={Q[l,m,0.2\linewidth]Q[l,m,0.3\linewidth]Q[l,m,0.2\linewidth]Q[l,m,0.3\linewidth]},
  column{1}={bg=lightgray, font=\bfseries},
  column{2}={bg=lightblue},
  column{3}={bg=lightgreen},
  column{4}={bg=headergray},
  hlines,
  vlines,
  cells={valign=m},
}
\textbf{Paradigm} & \textbf{Core Mechanism} & \textbf{Primary Data Modality} & \textbf{Representative Systems} \\
Modality-Driven Analysis & Visual/multimodal perception and interpretation & Images, charts, video, audio & ChartLlama \citep{han2023chartllama}, ChartAssisstant \citep{meng2024chartassisstant}, The AI Scientist-v2 \citep{yamada2025ai}, PlotGen \citep{luo2024plotgen} \\
Tool-Augmented Analysis & Interaction with external APIs, software, or hardware & Domain-specific data formats & ChemCrow \citep{bran2023chemcrow}, Coscientist \citep{boiko2023autonomous}, Virtual Lab \citep{boeck2023multiagent} \\
Computation-Native Analysis & General code generation/execution, symbolic reasoning & Tabular data, text, numerical arrays & Data Interpreter \citep{hong2024data}, LLM-SR \citep{kamienny2024llmsr}, MOBLLM \citep{binbas2024autonomous} \\

\end{tblr}
}
\label{tab:result_analysis}
\end{table}
\vspace{-1.0em}

\subsection{Iterative Result Validation and Refinement}
Scientific discovery rarely concludes after a single iteration of experiments. The iterative result refinement and validation phase emerges as a crucial mechanism for enhancing scientific outcomes. This stage involves multiple iterative cycles, each focused on carefully reviewing the results from previous stages, identifying discrepancies or unexpected findings, and methodically refining experimental procedures, computational models, or analytical techniques. Activities such as debugging computational workflows to detect and correct errors and systematically verifying adherence to established experimental protocols are essential to this process. Each iteration provides valuable feedback that informs subsequent improvements, thereby progressively aligning experimental outcomes with original hypotheses and research objectives. Concurrently, validation rigorously compares results against established scientific knowledge, predefined benchmarks, and reproducibility standards, ensuring the reliability, accuracy, and integrity of the scientific conclusions drawn.

\begin{figure*}[t!] 
    \centering 
    \includegraphics[width=\textwidth]{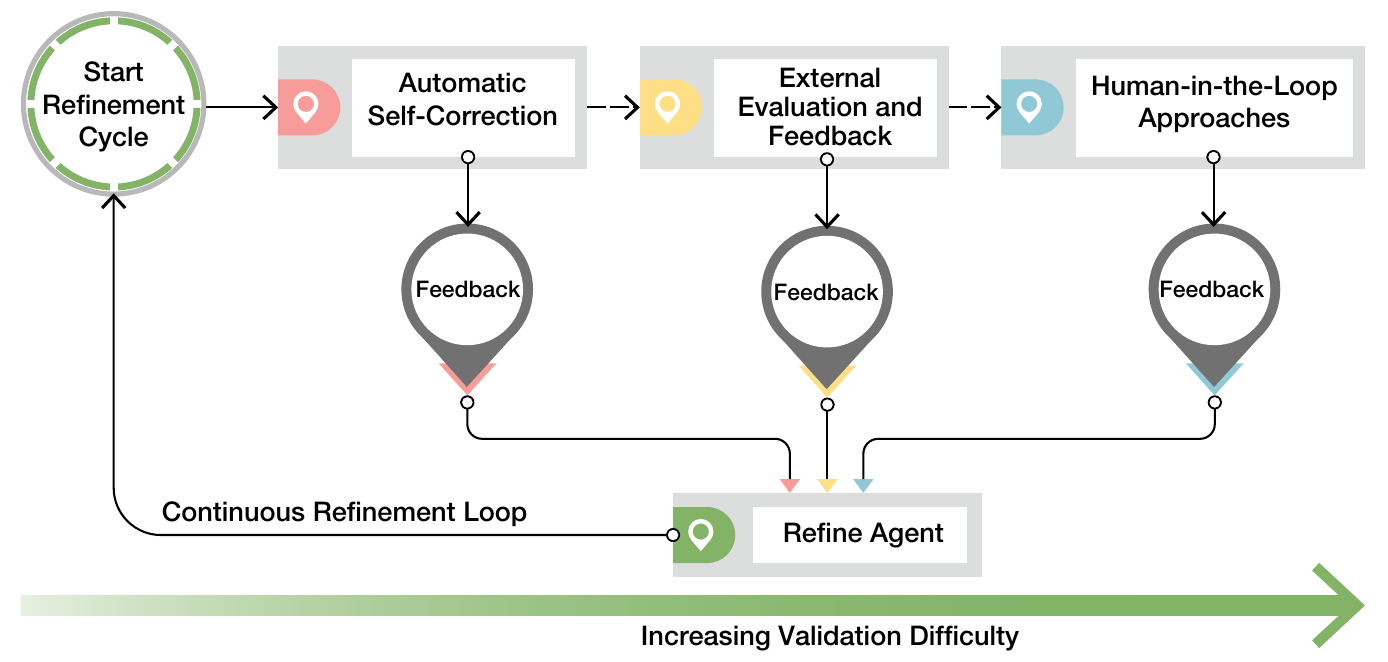} 
    \caption{A diagram of the scientific agent's continuous refinement cycle, which employs a progressive validation strategy. The process escalates through three stages as the difficulty of verifying the results increases. It begins with internal Automatic Self-Correction, moves to validation against tools and data via External Evaluation and Feedback, and for the most challenging cases, incorporates expert judgment through Human-in-the-Loop Approaches.} 
    \label{fig:refinement} 
\end{figure*}

Refinement within LLM-based systems typically uses three major strategies: automatic self-correction, external evaluation and feedback, and human-in-the-loop approaches (Figure \ref{fig:refinement}).

\textbf{Automatic Self-correction.}
Automatic self-correction employs introspective and self-debugging strategies, prompting LLMs to iteratively evaluate and refine their own outputs. This method typically involves the use of specialized prompts that encourage models to critique and revise their responses based on internal reasoning. For example, the Self-Refine framework \citep{madaan2023self} has shown significant improvements in reasoning tasks by enabling LLMs to critique and iteratively refine their responses. Additionally, frameworks such as those explored by \citet{chen2023teaching} highlight how iterative introspection significantly enhances code-generation accuracy. Within scientific LLM agents, ChemAgent uses dynamic memory retrieval to self-correct and refine reasoning based on prior experiences \citep{tang2025chemagentselfupdatinglibrarylarge}. 
SpatialAgent~\citep{wang2025spatialagent} relies on chain-of-thought reasoning and self-reflection to iteratively refine plans to achieve specific goals.
Team of AI-made Scientists (TAIS)~\citep{liu2024toward} uses multiple LLMs as distinct agents that provide feedback to one another, collaboratively refining and improving the overall results. BioInformatics Agent~\citep{xin2024bioinformatics} first summarizes and analyzes the results from the initial run, compares them with the user's expected outcomes, and then modifies the entire workflow to better align with the user's objectives. Recent works incorporate self-verification and self-correction mechanisms into reinforcement learning training to improve the robustness and reasoning capabilities of large language models \cite{zhang2025critique, jiang2025pag, liu2025trust}.
However, self-correction methods alone are often insufficient due to inherent limitations in LLM self-assessment capabilities~\citep{huang2023large}. 

\textbf{External Evaluation and Feedback.}
External evaluation and feedback strategies use tools and automated metrics external to the LLM to objectively validate and refine the model's outputs. These external methods include computational assessments, fact-checking tools, or specialized domain-specific tools to guide iterative improvement.  For instance, the CRITIC framework~\citep{gou2023critic} systematically validates LLM outputs using external fact-checking, such as a search engine or a knowledge base. In scientific applications, OSDA Agent~\citep{hu2025osda} uses computational chemistry tools to evaluate and iteratively refine molecular structure proposals. Similarly, MAPPS~\citep{zhou2025greaterautonomymaterialsdiscovery} integrates feedback from scientific computational tools and scientist insights to optimize experimental workflows and enhance discovery outcomes. Besides, the AI co-scientist~\citep{gottweis2025aicoscientist} employs an external multi-agent critique approach, facilitating peer-review-style evaluations that iteratively improve scientific hypotheses. BioDiscoveryAgent~\citep{roohani2024biodiscoveryagent} incorporates the phenotypic score from previous rounds in the prompt to help generate a better set of genes to experimentally perturb. SAMPLE~\citep{rapp2024self} connects with gene assembly, protein expression, and biochemical analysis systems, feeding the results back to refine the agent’s understanding in an iterative loop. Biomni~\citep{huang2025biomni} iteratively refines its reasoning using feedback from tools and code execution results.

\textbf{Human-in-the-Loop.}
Human-in-the-loop methods integrate direct human expertise and oversight into the LLM refinement process. These methods often involve structured iterative human evaluations, ensuring that outputs align closely with expert standards and real-world applicability. General research has extensively used reinforcement learning from human feedback (RLHF) to guide models toward behaviors reflecting expert evaluations. Within scientific contexts, AI co-scientist~\citep{gottweis2025aicoscientist} explicitly integrates human expertise within its iterative "generate, debate, and evolve" framework, ensuring robust scientific outcomes. ResearchAgent~\citep{baek2024researchagent} similarly uses human-aligned reviewing agents to simulate structured peer review, significantly enhancing the feasibility and novelty of scientific research proposals. MAPPS~\citep{zhou2025greaterautonomymaterialsdiscovery} relies on human scientists to verify the generated workflow to ensure seamless materials discovery, leading to more robust results.

\section{Domain-Specific Scientific Agents}
A new generation of scientific agents is being developed for fields as diverse as biology, chemistry, and materials science. By integrating and using advanced domain-specific tools, these agents are beginning to tackle challenging discovery tasks such as designing novel molecules, predicting protein structures, and discovering new materials with unprecedented speed.

In genomics, tools such as the UCSC Genome Browser \citep{kuhn2013ucsc}, AlphaGenome \citep{avsec2025alphagenome}, and Evo2 \citep{brixi2025genome} provide powerful resources for genome annotation, regulatory function prediction, and large-scale genome modeling~\citep{avsec2021effective,nguyen2024sequence,su2025learning,su2025language}.
In single cell biology, Seurat \citep{hao2021integrated} and Scanpy \citep{wolf2018scanpy} analyze large datasets. 
For proteomics, the AlphaFold family of models \citep{jumper2021highly,abramson2024accurate} and ESM series \citep{lin2023evolutionary,hayes2025simulating} predict protein structures and are increasingly applied in protein design workflows, while Rosetta \citep{leaver2011rosetta,watson2023novo,dauparas2022robust} serves as a versatile library for protein modeling and design and has given rise to some of the most powerful deep learning models in the field.
In chemistry, RDKit \citep{landrum2006rdkit} processes molecular data, and Q-Chem \citep{shao2006advances} and Psi4 \citep{smith2020psi4} perform quantum chemistry calculations. 
In materials science, ASE \citep{larsen2017atomic}, VASP \citep{kresse1996efficiency}, and LAMMPS \citep{plimpton1995fast,LAMMPS} simulate  molecular and materials behavior. Recently, machine learning is advancing molecular and materials simulation through Machine Learning Interatomic Potentials (MLIPs) and newer methods that directly predict quantum Hamiltonian matrices~\citep{yu2023efficient,yu2023qh9,yu2025QHNetV2}. In physics, tools like OpenFOAM \citep{OpenFOAM} for fluid dynamics and CIGALE \citep{noll2009cigale} for galaxy evolution enable complex simulations. This diverse suite of tools highlights how specialized computational resources are driving scientific discovery forward.

\textbf{Genomics.} 
The biological sciences have seen explosive growth in autonomous agent development. 
SpatialAgent~\citep{wang2025spatialagent} runs end-to-end spatial biology studies at scale, autonomously processing 2M+ cells across spatial transcriptomics and MERFISH pipelines to annotate tissue niches and map cell–cell interactions, demonstrating parity or gains against automated baselines while handling full projects with minimal human input. 
Team of AI-made Scientists (TAIS)~\citep{liu2024toward} is a framework that simulates researchers' works for genomics by assigning LLM agents specialized roles, including agents to plan the analysis, handle data selection and preprocessing, conduct biomedical research and analytical studies, analyze data statistics and interpret results, and review the quality of codes. Working together, these agents analyze gene expression datasets to identify disease-predictive genes, streamlining the pipeline of gene discovery (from raw data to insights) without human intervention.
CRISPR-GPT~\citep{huang2025crispr} automates gene-editing experiment design—from system and gRNA selection to delivery and validation protocols—and reports successful real-world use translating agent-generated designs into wet-lab execution. BioDiscoveryAgent~\citep{roohani2024biodiscoveryagent} chooses genes to perturb to achieve a target phenotype, outperforming trained Bayesian-optimization baselines by +21$\%$ on average (and +46$\%$ on non-essential genes) across six datasets, and doubling the hit-rate for multi-gene combinations over random. PerTurboAgent~\citep{hao2025perturboagent} plans iterative Perturb-seq campaigns, predicting next-round gene panels and improving panel quality over static or heuristic baselines in retrospective/simulation studies, thereby shortening cycles in pooled genetic screens. BioAgents~\citep{mehandru2025bioagents} delivers locally runnable, privacy-preserving bioinformatics workflows using multi-agent SLMs with RAG, achieving near-expert performance on conceptual genomics tasks while enabling on-premises customization with proprietary data. 
Single-cell genomics has also benefited greatly from multiple agent systems. BioInformatics Agent (BIA)~\citep{xin2024bioinformatics} runs end-to-end scRNA-seq analyses (dimensionality reduction, clustering, DE, enrichment), producing ready-to-use figures and reports with minimal user intervention. scBaseCount~\citep{youngblut2025scbasecount} uses an agentic, hierarchical workflow to continuously mine, standardize, and expand a single-cell compendium now exceeding 230 million cells across 21 organisms and 72 tissues, giving downstream agents a harmonized data backbone. CellAgent~\citep{xiao2024cellagent} automates scRNA-seq (and spatial) pipelines end-to-end via planner–executor–evaluator roles, delivering high-quality analyses without human steps and exportable narratives/plots for downstream interpretation. CompBioAgent~\citep{zhang2025compbioagent} provides natural-language exploration of scRNA-seq cohorts, turning user queries into JSON plans and returning targeted summaries and visualizations that lower the barrier for non-computational users. scAgent~\citep{mao2025scagent} targets universal cell-type annotation and novel-type discovery; across 160 cell types in 35 tissues it reports state-of-the-art accuracy and generalization, including data-efficient extension to unseen types. 
CASSIA~\citep{xie2024cassia} is a multi-agent annotator that produces interpretable cell-type calls with full reports and improves low-quality annotations by retrieving external evidence and consolidating tool outputs into audit-ready documentation. 
GeneAgent~\citep{wang2025geneagent} performs gene‑set analysis and queries expert‑curated biological databases to cross‑check its claims, reducing hallucinations and producing auditable rationales for functional descriptions.
STAgent~\citep{lin2025stagent} couples perception, dynamic code generation, tool use, and literature grounding to deliver end‑to‑end spatial transcriptomics analyses and structured reports with minimal human input.
MRAgent~\citep{xu2025mragent} automates Mendelian randomization studies end‑to‑end by triaging literature for exposure–outcome pairs, selecting GWAS, executing MR pipelines, and producing standardized reports.

\textbf{Protein.} The protein engineering domain has witnessed significant advances.
SAMPLE~\citep{rapp2024self} delivers a full closed-loop, self-driving protein engineering pipeline that experimentally discovers thermostable enzyme variants, identifying GH1 hydrolases with $\geq$12 °C higher stability than the starting sequences via autonomous design–build–test cycles in a robotic lab, rather than manual mutagenesis campaigns.
The Virtual Lab~\citep{swanson2024virtual} pushes from computation to wet-lab outcomes by designing 92 novel SARS-CoV-2 nanobodies and experimentally validating expression and binding across variants (with >90$\%$ expression in tested constructs), demonstrating that multi-agent pipelines can translate in-silico designs into viable binders.
ProtAgents~\citep{ghafarollahi2024protagents} demonstrates de novo proteins tailored to mechanical targets, generating sequences that meet target vibrational-frequency profiles and structural specifications validated through structure prediction and physics-aware analyses—showing agentic design can hit property constraints, not just folds. Sparks~\citep{ghafarollahi2025sparks} shows agentic discovery can move from instances to general principles, autonomously uncovering two protein-design rules: a length-dependent crossover where $\beta$-biased peptides surpass $\alpha$-helical ones in unfolding force beyond $\sim$80 residues, and a chain-length/secondary-structure stability map with a high-variance “frustration zone” for $\alpha/\beta$ folds. VibeGen~\citep{ni2025agentic} extends this to dynamics-aware design, where a designer–predictor duo produces de novo proteins whose all-atom MD trajectories reproduce prescribed normal-mode amplitudes, expanding beyond evolutionary templates toward motion-tuned biomolecules. AutoProteinEngine (AutoPE)~\citep{liu2024autoproteinengine} lowers the barrier for biologists by automating model choice, HPO, and data handling; on two real protein-engineering tasks it shows substantial accuracy gains over zero-shot and manual fine-tuning, turning advanced DL pipelines into practical, conversational workflows. ProtChat~\citep{huang2024protchat} further streamlines protein analysis tasks into a multi-agent workflow by combining a high-level generalist LLM (e.g. GPT-4o) for task-planning, chatting and visualization, with specialized PLMs (e.g. ESM) for protein understanding.
PRIME~\citep{zhou2025prime} interprets high‑level goals and dynamically synthesizes computational workflows for protein engineering by reasoning over a curated library of 65+ validated protein tools, constructing customized pipelines end‑to‑end.

\textbf{Medicine.} 
A recent groundbreaking study is Biomni~\citep{huang2025biomni}, which functions as a general-purpose biomedical agent that \emph{actually completes} end-to-end analyses—causal gene prioritization, drug repurposing, rare-disease workups, and multi-omics integration—by driving 150+ tools, 105 software packages, and 59 databases, returning ranked candidates with traceable evidence.
AI co-scientist~\citep{gottweis2025towards} operationalizes hypothesis-driven research: a Gemini-2.0 multi-agent system that turns literature + model feedback (e.g., structure prediction) into novel, testable hypotheses and study proposals with iterative review. TxAgent~\citep{gao2025txagent} converts patient context into actionable therapy decisions, using 211 tools for up-to-date interaction checks, contraindications, and personalized regimen selection via multi-step tool-augmented reasoning. Beyond these, BioResearcher~\citep{luo2025intention} demonstrates dry-lab automation end-to-end—from goal $\rightarrow$ literature synthesis $\rightarrow$ executable analysis/protocol drafts $\rightarrow$ reviewer-style critiques—showing measurable gains on complex research objectives without manual “glue code.” STELLA~\citep{jin2025stella} targets the evidence-curation bottleneck in biomedicine with a self-evolving, multi-agent literature analysis workflow that structures large corpora into machine-navigable knowledge to support hypothesis generation/validation. In the domain of drug discovery, CLADD~\citep{lee2025rag} coordinates collaborative agents with RAG to design/dock/triage while ingesting heterogeneous biochemical knowledge without fine-tuning, improving task performance over general LLMs and classical DL baselines.
DrugAgent~\citep{liu2024drugagent} automates end-to-end ML programming for ADMET/repurposing tasks (data acquisition $\rightarrow$ training $\rightarrow$ evaluation), reporting strong case-study metrics (e.g., PAMPA absorption $F1\approx 0.92$).
PharmAgents~\citep{gao2025pharmagents} builds a “virtual pharma” where specialist agents span screening $\rightarrow$ modeling $\rightarrow$ triage, outputting prioritized leads with assay-ready metadata.
LIDDiA~\citep{averly2025liddia} is an autonomous, language-driven discovery agent that generates molecules meeting pharmaceutical criteria on many targets and surfaces promising EGFR candidates, emphasizing low-cost adaptability.
AgentMD~\citep{jin2024agentmd} automatically selects and executes from 2,164 curated clinical calculators (RiskCalcs) to return risk estimates with formula provenance, substantially outperforming strong prompting baselines on RiskQA.
MedAgents~\citep{tang2023medagents} delivers multi-specialist deliberation in zero-shot settings to reach consensus diagnoses/plans across standard medical QA suites.
ClinicalGPT~\citep{wang2023clinicalgpt} is a domain LLM for clinical tasks; EHR-integrated agentic examples include EHRAgent, which auto-generates/executes code against EHR data to answer complex patient queries and compute scores.
BehaveAgent~\citep{aljovic2025autonomous} provides turn-key cross-species behavior analysis from raw video—planning the analysis, tracking/pose estimation, sequence labeling, and report generation—without retraining across paradigms.

\textbf{Chemistry.} Chemistry has emerged as a particularly fertile ground for autonomous agents. For instance, the agent Coscientist~\citep{boiko2023autonomous} tackled the overarching problem of autonomously designing and executing complex physical experiments from high-level prompts, demonstrating its capability by successfully performing Nobel Prize-winning palladium-catalyzed cross-coupling reactions in under 4 minutes. To broaden the utility of LLMs in chemistry, ChemCrow~\citep{m2024augmenting} addressed the challenge of equipping them with specialized knowledge by integrating 18 expert-designed tools, successfully performing autonomous planning and multi-tool analysis across organic synthesis, drug discovery, and materials science. For managing highly complex workflows, ChemAgents~\citep{song2025multiagent} solved the need for distributed expertise by deploying a hierarchical multi-agent system where a manager agent coordinates specialized agents for tasks ranging from literature analysis to robotic operation. Bridging linguistic reasoning with quantum mechanics, ChemReasoner~\citep{sprueill2024chemreasoner} focused on the difficult task of catalyst discovery by creating a synergistic loop between LLM-driven hypothesis generation and rapid feedback from DFT simulations, thereby accelerating the materials design cycle. To improve accessibility, CACTUS~\citep{mcnaughton2024cactus} solved the user-interface problem for complex computational chemistry tools, acting as an intelligent conversational assistant that accurately translates natural language questions into the proper tool calls for property prediction, similarity searches, and toxicity estimation. In the domain of experimental optimization, Chemist-X~\citep{chen2023chemist} focused on automating the tedious process of refining reaction conditions, achieving a fully automated, closed-loop system that uses retrieval-augmented generation to propose conditions and then directs a robotic platform for wet-lab validation. In computational chemistry, El Agente Q~\citep{zou2025agente} tackled the manual, error-prone steps in computational chemistry, demonstrating how its cognitive architecture could autonomously handle an entire workflow—from file preparation to cluster submission and result parsing—based on a simple natural language prompt. To empower bench chemists, LLM-RDF~\citep{ruan2024automatic} aimed to provide a fully automated synthesis workflow for users without coding expertise, successfully enabling an end-to-end process from literature search to product purification through its conversational, six-agent framework. In the pharmaceutical space, FROGENT~\citep{pan2025frogent} addressed the complex, end-to-end process of drug design, showing superior performance on multi-step discovery benchmarks by integrating diverse biochemical databases and predictive models into a unified framework. LARC~\citep{baker2025larc} solved the critical challenge of applying practical, real-world constraints to retrosynthesis planning, employing an Agent-as-a-Judge mechanism to generate more feasible routes and achieving a near-human-level success rate on diverse tasks. FMG \citep{sun2025foundation} demonstrates expert-level molecular design can be achieved by adopting graph representations and rendering them as images, highlighting multiple modalities can enable deeper understanding. Finally, addressing the core of scientific inquiry, MOOSE-Chem~\citep{yang2024moose} aimed to automate the creative process of hypothesis generation by mimicking human cognition, successfully rediscovering the core innovations from recent high-impact chemistry papers without any prior knowledge of their content.

\textbf{Materials.} The application of scientific agents in materials science has led to innovative solutions for long-standing research challenges, spanning from discovery to synthesis and analysis. The A-Lab ~\citep{szymanski2023autonomous} of Lawrence Berkeley National Laboratory represents a fully autonomous materials synthesis facility. Using three robotic arms, box furnaces, and X-ray diffractometers, it synthesized 41 novel compounds from 58 targets over 17 days of continuous operation, achieving a 71\% success rate. ChatMOF~\citep{kang2024chatmof} addresses the complex scientific problem of discovering novel Metal-Organic Frameworks (MOFs), a process that requires navigating vast chemical databases, predicting properties of hypothetical structures, and generating new candidates. Its primary scientific contribution is a unified, autonomous workflow that orchestrates these disparate tools, successfully translating high-level goals into a concrete series of actions to accelerate the identification of promising materials for applications like gas storage. Similarly, the ``ChatGPT Research Group''~\citep{zheng2023chatgpt} tackles the resource-intensive, iterative process of optimizing experimental synthesis conditions for advanced materials. Its achievement is a collaborative multi-agent system that mimics a human research team, dividing labor among specialized LLM agents to significantly accelerate the experimental optimization cycle and make the discovery-to-production pipeline more systematic. In the computational domain, MDAgent~\citep{shi2025fine} confronts the efficiency bottleneck in Molecular Dynamics (MD) simulations, which are critical for understanding material behavior but require significant user expertise. It automates the entire MD workflow from code generation to execution, achieving a notable scientific result by reducing the total task time for thermodynamic calculations by over 40\% and lowering the barrier to entry for performing these complex simulations. To address the limitations of general-purpose LLMs, HoneyComb~\citep{zhang2024honeycomb} solves the problem of their lack of deep, domain-specific knowledge and reliability in scientific computations. It contributes a robust agent framework grounded with a curated materials knowledge base and a hub of validated scientific tools, resulting in significantly higher accuracy in both reasoning and computation. Addressing the grand challenge of inverse design, LLMatDesign~\citep{jia2024llmatdesign} introduces a framework for autonomous materials discovery, particularly in low-data regimes. Its key scientific achievement is a self-reflection mechanism that allows the agent to learn from computational outcomes and adapt its strategy, enabling the effective discovery of materials with specific, user-defined properties. Meanwhile, MatAgent~\citep{bazgir2025matagent} targets the fragmentation and lack of reproducibility in typical materials discovery workflows. It provides an integrated, human-in-the-loop multi-agent framework that streamlines the research process, fostering an AI-guided, adaptive laboratory workflow that enhances both the speed and reproducibility of materials research. For managing multi-faceted research projects, MatSciAgent~\citep{chaudhari2025modular} solves the challenge of orchestrating diverse computational tasks. It achieves this with a modular architecture where a master agent interprets high-level goals and delegates tasks to specialized sub-agents, creating an extensible and robust system for complex research questions. To better integrate human expertise, MatPilot~\citep{ni2024matpilot} demonstrates a powerful human-machine collaborative framework. It resolves the challenge of seamlessly combining human intuition with AI's computational power by using multi-agent LLMs for hypothesis generation while allowing human experts to guide the overall strategy, enabling a continuous learning loop. Materials Laws Multi-Agent Framework~\citep{hu2024multi} tackles the fundamental scientific goal of distilling complex data into simple, interpretable physical laws. Its major scientific result was the successful use of LLM agents to perform symbolic regression, deriving a low-complexity, highly accurate formula for predicting glass-forming ability in metallic glasses and showcasing the potential of AI to automate scientific law discovery.

\textbf{Physics.} Scientific agents are increasingly being applied to solve complex problems in physics and engineering, demonstrating capabilities ranging from controlling physical hardware to automating sophisticated simulations. In quantum computing, the k-agents framework~\citep{cao2024agents} addresses the scientific challenge of automating quantum experiment design and execution, a process that typically requires deep human expertise. Its significant scientific achievement was the creation of a fully autonomous laboratory where LLM agents successfully planned and executed experiments on superconducting quantum processors to produce entangled states at a performance level equivalent to human experts. In materials physics, AtomAgents~\citep{ghafarollahi2024atomagents} tackles the complex, multi-factorial problem of designing new alloys with specific properties. It contributes a physics-aware, multi-agent framework that successfully integrates knowledge retrieval with physics-based simulations in an iterative loop, enabling the system to autonomously propose, simulate, and refine alloy compositions to meet performance targets. In astrophysics, Mephisto~\citep{sun2024interpreting} solves the complex inverse problem of interpreting multi-band galaxy observations to determine their physical properties. Its key result is a multi-agent framework that automates this by calling the CIGALE astrophysics code as a tool, successfully proposing and testing hypotheses against observational data in an iterative reasoning loop. In quantum instrumentation, QCopilot~\citep{sha2025llm} confronts the challenge of designing and diagnosing highly sensitive quantum sensors, a process that involves time-consuming manual parameter tuning. It achieved a remarkable ~100-fold speedup over manual procedures by autonomously performing modeling, optimization, and anomaly detection in atom cooling experiments. In computational fluid dynamics (CFD), OpenFOAMGPT~\citep{pandey2025openfoamgpt} solves the problem of the steep learning curve and tedious setup process associated with complex simulation software like OpenFOAM. It successfully automates the entire workflow, from case setup to iterative correction, significantly lowering the barrier to entry for advanced CFD. MetaOpenFOAM~\citep{chen2024metaopenfoam} further addresses the need for robust and generalized CFD automation by using a multi-agent system to decompose complex natural language instructions, achieving strong performance across a diverse range of flow simulations. Finally, to address the critical need for interpretability in scientific AI, the AI-Scientist Framework~\citep{xu2025advancing} introduces a multi-agent system that structures LLM outputs into transparent, executable models. Its main contribution is enhancing systematic validation and human-AI collaboration by making the agent's reasoning process physically grounded and verifiable.

\textbf{Other Science and Engineering.} Beyond the foundational scientific disciplines, scientific agents are driving significant advancements in a range of other scientific and engineering fields. The Autonomous GIS Agent~\citep{ning2025autonomous} and its successor GIS Copilot~\citep{akinboyewa2025gis} address the accessibility barrier in geospatial science, where answering spatial questions often requires specialized programming skills. They successfully automated workflows from data retrieval to the generation of maps and statistics by translating natural language requests into executable programs, making advanced geospatial analysis more accessible. For complex engineering systems, the domain-specific ReAct agent for gas turbines~\citep{song2024domain} tackles the challenge of modeling and analyzing the thermodynamics of multi-component systems. It successfully integrated expert knowledge with predefined tools to perform iterative gas path analysis, demonstrating a viable path for AI-driven diagnostics in power engineering. LP-COMDA~\citep{liu2024physics} extends agent applications to power electronics, solving the challenge of designing complex and error-prone modulation strategies for power converters. It successfully automated this process with a physics-informed planner, accelerating design time by over 30x while reducing errors by over 60\%.

\section{Discussions}

\subsection{From LLM Reasoning to Agentic Reasoning via Reinforcement Learning}

The advent of Large Language Models (LLMs) has brought the promise of automating complex tasks, but Reinforcement Learning (RL) plays a crucial role in transitioning from passive text generation to active decision-making and discovery. While Supervised Fine-Tuning (SFT) can enhance domain knowledge and change model behaviors, it is fundamentally a form of imitation learning, limiting a model's ability to generalize to new domains \citep{chu2025sft} and explore more optimal or diverse problem-solving strategies. Reinforcement Learning, particularly from verifiable rewards, provides a fundamental solution~\citep{parashar2025curriculum,su2025iterative,li2024derivative,li2025dynamic} to elicit generalizable reasoning ability from the base policy model. Work exemplified by DeepSeek-R1~\citep{guo2025deepseek} demonstrates that models can learn complex reasoning behaviors, including self-verification and self-correction, by being rewarded solely from final answer correctness. These emergent capabilities reveal RL's potential for transforming LLMs from passive knowledge repositories into active problem solvers.

The agentic RL paradigm extends model behaviors from pure reasoning to actionable interactions with environments, where agents learn to dynamically interleave reasoning and tool execution. Through RL, agents optimize not just what to think, but when and how to act.
Code interpreters enhance mathematical abilities~\citep{feng2025retool,mai2025agent} of agents, with agents learning through RL to optimize reasoning trajectories via multi-turn real-time code execution. Search engines enable agents to acquire external knowledge and up-to-date information, where RL helps agents generate effective search queries as part of their reasoning process \citep{jin2025searchr1,chen2025research}.
Beyond augmenting capabilities, agentic RL also enables models to operate in complex, dynamic environments. Web browsing agents \citep{wei2025webagentr1,qi2025webrl} and GUI agents \citep{qin2025ui,agashe2024agent} position models in human-like scenarios, requiring them to interact with environments where states change in response to their actions. Agentic RL foster the model capability in handling continuous perception, decision-making, and action execution in response to environmental feedback. Beyond a single tool, RL also trains agents to coordinate multiple tools concurrently to solve harder tasks like deep research~\citep{geng2025webwatcher,li2025webthinker}.

Applying the successful Agentic RL methods to scientific discovery is a promising way for realizing an AI-driven scientific revolution. However, the unique nature of the scientific domain presents challenges that far exceed those in existing work.

\textbf{The Environment Problem.} The environment for a Science Agent is an unavoidably heterogeneous hybrid. The system parses outputs from diverse digital tools with non-standard interfaces (e.g., command-line software, database APIs, graphical user interfaces) and interacts directly with physical laboratory equipment such as robotic arms and chemical synthesizers. The introduction of the physical world brings with it latency, noise, irreversible actions, and the risk of hardware damage—challenges that purely digital agents have never faced.

\textbf{The Action Problem.} The action space for a Science Agent, however, is enormous and non-standard, as the actions to control a mass spectrometer are entirely different from those to call a data analysis library. More importantly, an advanced Science Agent is expected not only to use tools, but also to create new ones by writing code. This expands the action space from a finite set to the near-infinite space of all possible programs, posing a massive challenge for exploration-based RL. While the Agent RL Scaling Law study reveals that performance scales with the number of interactions, the astronomical number of interactions required to effectively explore the explosive action space of a Science Agent is practically infeasible.

\textbf{The Observation Problem.} A Science Agent needs to process mixed-modality observations, including experimental images, mass spectrometry data, simulation curves, structured data like SMILES chemical formulas, and unstructured text from scientific papers. Furthermore, research projects have long durations, requiring the agent to possess long-term memory capabilities to connect a serendipitous finding from months ago to a current problem. This far exceeds the capacity of existing agents that primarily rely on short-term context for decision-making and places extreme demands on the model's memory and retrieval systems.

\textbf{The Reward Problem.} All successful Agent RL works rely on a clear, sparse, but definitive final reward signal, such as whether a math problem is correct, a web task is completed, or a question is answered correctly \citep{jin2025searchr1,wei2025webagentr1,chen2025rm}. This paradigm almost completely breaks down in scientific discovery. The reward is highly sparse, as a true discovery can take years, and the outcome of a single experiment does not serve as an effective reward signal~\citep{uehara2025inference,uehara2025reward_paper}. The objective of success is also ambiguous. Designing a reward function that can measure novelty, impact, or reproducibility is a major, unresolved problem. Although the self-evolving curriculum proposed in WebRL offers a promising direction by generating easier sub-tasks from failures, it is still confined to tasks with clear success criteria. For open-ended scientific exploration, defining and calculating the reward remains the most central obstacle on the path to a general-purpose Science Agent.

\subsection{The Leap from LLMs to Agents in Autonomous Scientific Discovery}

\begin{figure}[t]
    \centering
    \includegraphics[width=0.7\textwidth]{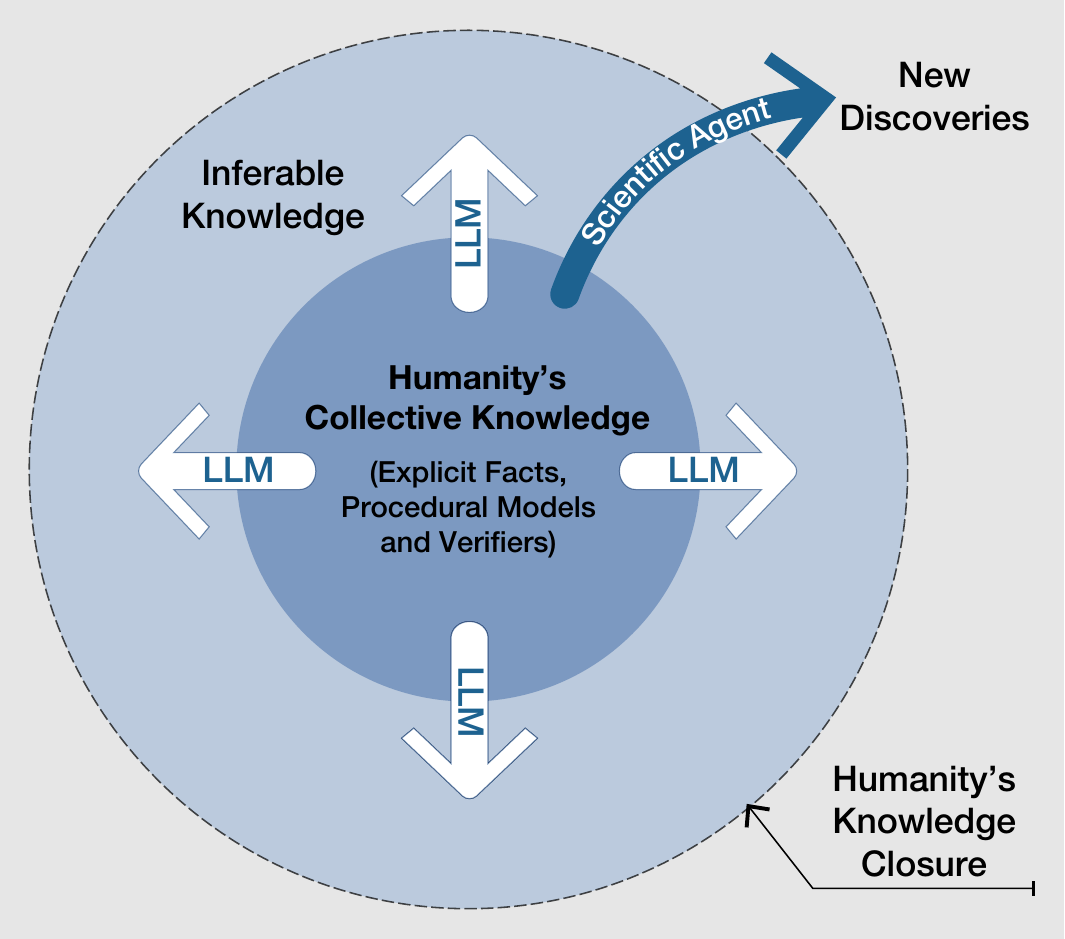}
    \caption{A conceptual model illustrating the limitation of a Large Language Model (LLM) compared to the potential of a Scientific Agent. An LLM, acting purely as a reasoning engine, is fundamentally confined within Humanity's Knowledge Closure. It excels at processing and making connections within the space of existing and Inferable Knowledge but struggles to generate truly novel insights that lie outside this boundary. In contrast, a Scientific Agent is designed to overcome this limitation. By actively interacting with the world—designing experiments, executing actions, and employing exploration strategies—it can break through the knowledge closure to produce genuinely New Discoveries, thereby expanding the frontier of human understanding. This discussion was motivated by and adapted from \citet{Rao:knowledge}, and the figure is expanded and recreated with permission.}
    \label{fig:knowledge}
\end{figure}

LLMs are primarily seen as powerful reasoning engines, like our human brain. Trained on vast amounts of data, they excel at processing, analyzing, and verifying existing human knowledge. However, even if an LLM could propose a novel, unconventional hypothesis, it lacks the ability to physically interact with the world to verify its own ideas. It is not able to design experiments, conduct tests, or observe real-world results, which are essential steps for turning a hypothesis into a validated discovery. This limitation is visually depicted in the Figure \ref{fig:knowledge}, where the LLM is trapped inside the Humanity's Knowledge Closure. It performs efficient reasoning within the existing circle of knowledge but struggles to cross this boundary to reach New Discoveries.

To overcome this challenge, the LLM needs to evolve from a passive reasoning engine into an active scientific agent. As \citet{Rao:knowledge} emphasizes, human can interact with the world and explore new knowledge, and a complete scientific agent should also possess this ability. From an information-theoretic perspective, as detailed in Section \ref{sec:Information}, the agent's internal entropy reduction is fundamentally dependent on irreversible interactions with the physical world. The necessary constraining information cannot be generated internally but must be obtained from real-world processes to truly reduce the hypothesis space.

A scientific agent can not only process data but also design and execute experiments, actively exploring unknown domains. They can acquire new data through trial and error, observation, and hands-on actions, much like a human scientist, rather than solely relying on existing information. A scientific agent is no longer a passive recipient of information; it can translate its internal knowledge (explicit facts, procedural models, etc.) into concrete actions in the real world. This action-feedback loop allows the agent to continuously break through the boundaries of inferable knowledge,  gradually expanding the circle of human knowledge.

In this way, an LLM-driven scientific agent will no longer be merely an extension of human thought but can become an independent, exploratory pioneer. It can assist humans in efficiently exploring the frontiers of inferable knowledge and may even, through a chance deviation, help humanity discover entirely new insights and groundbreaking discoveries that truly lie beyond the current knowledge closure.

\subsection{Role of Serendipity in Discovery}

An essential aspect of scientific discovery is the inherent unpredictability and the role that chance and unexpected findings play in significant breakthroughs. Historical milestones, such as the discovery of penicillin, cosmic microwave background radiation, and graphene, highlight how serendipitous findings are pivotal in driving scientific advancement. This phenomenon underscores a crucial gap between human-driven scientific inquiry and current capabilities of LLM-based science agents.

LLMs are trained to maximize likelihood based on extensive historical data, inherently optimizing towards predictable and probable outcomes. Such a training objective inadvertently constrains the models' abilities to venture into the realms of randomness and serendipity that frequently yield novel insights in human science. Major historical scientific breakthroughs, such as the discovery of penicillin~\citep{Fleming1929}, cosmic microwave background radiation~\citep{Penzias1965}, and graphene~\citep{Novoselov2004}, often originated from chance or unexpected observations that break free from the confines of pure logical deduction. Consequently, purely likelihood-driven agents might overlook unconventional hypotheses or unexpected patterns that do not align closely with existing data distributions.
To bridge this gap, specific mechanisms or design principles could be integrated into LLM-based science agents to better accommodate chance and unexpected discoveries. For instance, stochastic exploration strategies could be employed to occasionally deviate from the highest probability outputs, thereby simulating the conditions under which human scientists encounter serendipitous results. 

\section{Summary and Outlook}

The emergence of LLM-based agents represents a substantial step forward in the automation and acceleration of scientific discovery. By addressing critical limitations inherent to traditional human-driven approaches and reinforcement learning-based computational agents, LLM-driven methods provide a unified and highly flexible framework capable of operating seamlessly across human intent, natural language, computer language, and physical information. This unified capacity significantly enhances the adaptability, scalability, and generalizability of scientific agents, enabling dynamic engagement across all stages of the discovery lifecycle.

As scientific inquiry continues to confront increasingly complex challenges and larger volumes of data, the continued development and refinement of LLM-based scientific agents become essential. Future research should focus on overcoming existing limitations, such as better understanding physical laws, integrating robust mechanisms for using physical tools, and designing sophisticated interactions with experimental environments and physical instrumentation. Such advancements will further strengthen the capabilities of scientific agents, promoting deeper collaboration between humans and computational intelligence. Ultimately, the ongoing evolution of LLM-based agents holds the promise of revolutionizing scientific discovery, enabling unprecedented efficiency, adaptability, and creativity in generating novel, transformative insights across all scientific disciplines.

\section*{Acknowledgments}

The authors thank Ms. Xiaomeng Fu for the professional graphic design support. S.J. gratefully acknowledges support from the National Science Foundation under grants IIS-2243850, MOMS-2331036, and CNS-2328395; the Advanced Research Projects Agency for Health under grant 1AY1AX000053; and the National Institutes of Health under grant U01AG070112. Additional support to S.J. was provided by the Texas A\&M Institute of Data Science, the Truchard Family Endowed Chair, the Presidential Impact Fellowship, and the Chancellor EDGES Fellowship at Texas A\&M University. W.W. gratefully acknowledges support from the National Science Foundation under grants 2106859, 2119643, 2200274, 2202693, 2312501; the National Institute of Health under grants U54OD036472, U54HG012517, U24DK097771, OT2OD038003; the United States Department of Agriculture under grant 13434200; as well as support from Amazon, NEC, and Optum AI. X.Q. gratefully acknowledges support from the National Science Foundation under grants CMMI-2226908, DMR-2103842, and DMR-1753054; the Air Force Office of Scientific Research under grant FA9550-24-1-0207; and partial support by the donors of ACS Petroleum Research Fund
under grant \#65502-ND10. H.J. acknowledges support from the Molecule Maker Lab Institute: an AI research institute program supported by NSF under award No. 2019897 and No. 2034562. The views and conclusions contained herein are those of the authors and should not be interpreted as necessarily representing the official policies, either expressed or implied, of the U.S. Government. The U.S. Government is authorized to reproduce and distribute reprints for governmental purposes notwithstanding any copyright annotation therein. H.J. is also supported by DOE Center for Advanced Bioenergy and Bioproducts Innovation U.S. Department of Energy, Office of Science, Office of Biological and Environmental Research under Award Number DE-SC0018420.

\bibliography{example_paper,dive,AI4Sci}
\bibliographystyle{unsrtnat}

\end{document}